%% file: main.tex
\IfFileExists{arxiv/main.tex}{\def\arxivpath{arxiv/}\def\bytedancefontpath{}}{\def\arxivpath{}\def\bytedancefontpath{seed/}}
\let\projectpath\arxivpath

\edef\bytedanceassetpath{\arxivpath seed/}
\makeatletter
\edef\input@path{{\arxivpath}}
\makeatother

\documentclass[10pt]{bytedance}

\ifdefined\pdfsuppressptexinfo \pdfsuppressptexinfo=-1 \fi

\input{template.tex}
\input{preamble.tex}

\input{metadata.tex}

\abstract{%
  \begingroup
  \renewenvironment{abstract}{}{}%
  \input{sections/00_abstract}  \endgroup
}

\begin{document}
\vspace*{-8mm}
\maketitle

\input{content.tex}

\end{document}

%% file: template.tex
\IfFileExists{sfrm1000.pfb}{}{%
  }

\fancypagestyle{firststyle}{%
  \fancyhead[R]{%
    \vskip 1.8mm
    \raisebox{1.35mm}{\includegraphics[height=10.5mm]{\bytedanceassetpath shanghaitech.pdf}}%
    \hspace{3.5mm}%
    \raisebox{1.35mm}{\includegraphics[height=10.5mm]{\bytedanceassetpath bigai.png}}}%
  \fancyhead[L]{%
    \vskip 1.8mm
    \includegraphics[width=57.156mm,height=13.2mm]{\bytedanceassetpath bytedance.jpg}}%
}
\DeclareFontShape{T1}{bytesans}{b}{n}{<-> s * [1] \bytedancefontpath bytesans}{}
\DeclareFontShape{T1}{bytesans}{bx}{n}{<-> s * [1] \bytedancefontpath bytesans}{}
\DeclareFontShape{T1}{bytesans}{bx}{sc}{<-> s * [1] \bytedancefontpath bytesans}{}
\renewcommand{\authorformat}[2][]{\mbox{\authorfont\sffamily #2$^{#1}$}}
\renewcommand{\affiliation}[2][]{\addtolist[#1]{#2}{\affiliationlist}{\affiliationformat}{\qquad}}

\graphicspath{{\projectpath}{\arxivpath}}

%% file: preamble.tex
\usepackage[utf8]{inputenc}
\usepackage{latexsym}
\IfFileExists{inconsolata.sty}{\usepackage{inconsolata}}{}
\input{\arxivpath math_commands.tex}

\usepackage{amssymb}
\usepackage{url}
\usepackage{xurl}
\usepackage{algorithm}
\usepackage{algpseudocode}
\usepackage{tabularx}
\usepackage{enumitem}
\usepackage{adjustbox}

\AtBeginDocument{%
  \setlength{\parfillskip}{0pt plus 0.85\textwidth}%
  \setlength{\emergencystretch}{1.5em}}
\tcbset{every box/.style={before upper={\setlength{\parfillskip}{0pt plus 0.85\linewidth}}}}

\BeforeBeginEnvironment{tabular}{\begin{adjustbox}{max width=\linewidth}}
\AfterEndEnvironment{tabular}{\end{adjustbox}}

\usepackage{letltxmacro}
\usepackage{environ}
\makeatletter
\providecommand{\@LN}[2]{}
\def\arxiv@columnwidthtoken{\columnwidth}
\newif\ifarxiv@inpair

\newcommand{\arxivfigurewidth}[2]{\@namedef{arxiv@figw@#1}{#2}}
\LetLtxMacro{\arxiv@includegraphics}{\includegraphics}
\DeclareRobustCommand{\includegraphics}{%
  \@ifstar{\arxiv@includegraphics*}{\arxiv@includegraphics@}}
\newcommand{\arxiv@includegraphics@}[2][]{%
  \ifarxiv@inpair
    \arxiv@includegraphics[#1]{#2}%
  \else\ifcsname arxiv@figw@#2\endcsname
    \arxiv@includegraphics[width=\csname arxiv@figw@#2\endcsname]{#2}%
  \else
    \arxiv@includegraphics[#1]{#2}%
  \fi\fi}

\arxivfigurewidth{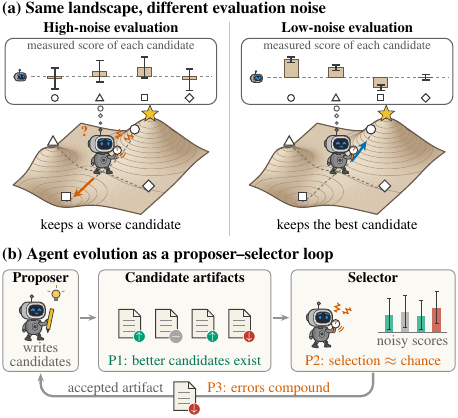}{0.6\linewidth}
\arxivfigurewidth{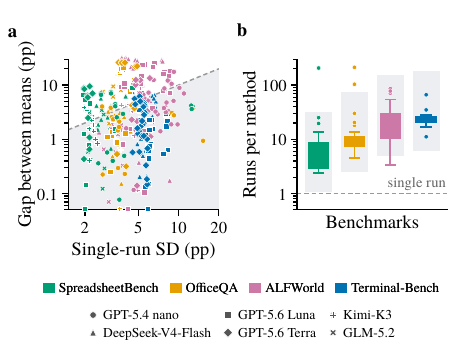}{0.62\linewidth}
\arxivfigurewidth{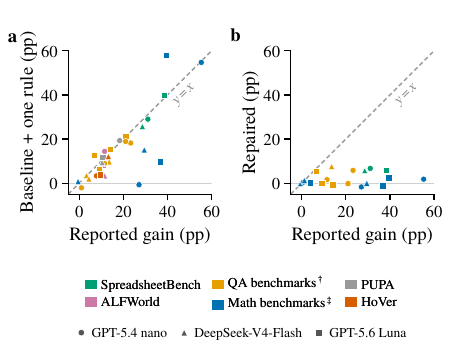}{0.62\linewidth}
\arxivfigurewidth{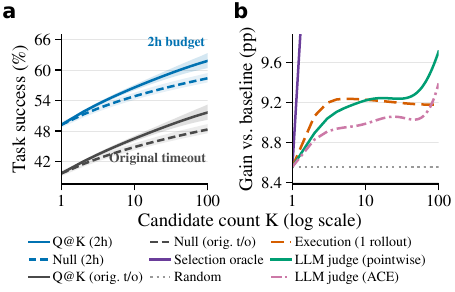}{0.65\linewidth}
\arxivfigurewidth{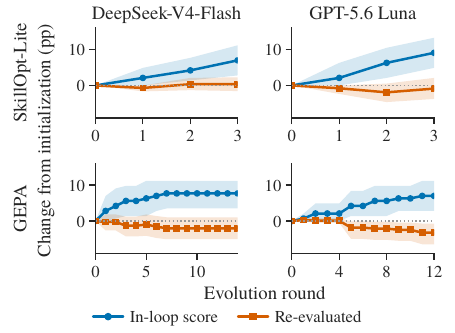}{0.62\linewidth}
\arxivfigurewidth{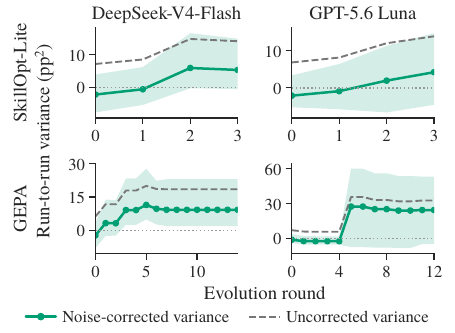}{0.62\linewidth}

\LetLtxMacro{\arxiv@resizebox}{\resizebox}
\DeclareRobustCommand{\resizebox}{%
  \@ifstar{\arxiv@resizebox*}{\arxiv@resizebox@}}
\newcommand{\arxiv@resizebox@}[3]{%
  \ifarxiv@inpair\def\arxiv@tmp{}\else\def\arxiv@tmp{#1}\fi
  \ifx\arxiv@tmp\arxiv@columnwidthtoken
    \expandafter\@firstoftwo
  \else
    \expandafter\@secondoftwo
  \fi
  {\adjustbox{max width=\linewidth}{#3}}%
  {\arxiv@resizebox{#1}{#2}{#3}}}

\newcommand{\arxivnarrowtablewidth}{0.6\linewidth}
\let\arxiv@tabularx\tabularx
\def\tabularx#1{%
  \ifarxiv@inpair\def\arxiv@tmp{}\else\def\arxiv@tmp{#1}\fi
  \ifx\arxiv@tmp\arxiv@columnwidthtoken
    \expandafter\@firstoftwo
  \else
    \expandafter\@secondoftwo
  \fi
  {\arxiv@tabularx{\arxivnarrowtablewidth}}%
  {\arxiv@tabularx{#1}}}

\newcommand{\arxivpairslotwidth}{0.485\linewidth}
\def\arxiv@pairlabels{}
\newcommand{\arxivpair}[2]{%
  \@namedef{arxiv@partner@#1}{#2}%
  \@namedef{arxiv@partner@#2}{#1}%
  \@namedef{arxiv@pairfirst@#1}{#1}\@namedef{arxiv@pairsecond@#1}{#2}%
  \@namedef{arxiv@pairfirst@#2}{#1}\@namedef{arxiv@pairsecond@#2}{#2}%
  \edef\arxiv@pairlabels{%
    \arxiv@pairlabels\ifx\arxiv@pairlabels\@empty\else,\fi#1,#2}}
\let\arxiv@figure\figure
\let\arxiv@endfigure\endfigure
\let\arxiv@table\table
\let\arxiv@endtable\endtable
\expandafter\let\csname arxiv@figure*\expandafter\endcsname\csname figure*\endcsname
\expandafter\let\csname arxiv@endfigure*\expandafter\endcsname\csname endfigure*\endcsname
\expandafter\let\csname arxiv@table*\expandafter\endcsname\csname table*\endcsname
\expandafter\let\csname arxiv@endtable*\expandafter\endcsname\csname endtable*\endcsname
\long\def\arxiv@getlabel#1\label#2#3\arxiv@nil{\def\arxiv@thislabel{#2}}
\newcommand{\arxiv@float}[2]{
  \expandafter\arxiv@getlabel\BODY\label{}\arxiv@nil
  \ifcsname arxiv@prefetch@\arxiv@thislabel\endcsname
    \ifarxiv@prereading
      \expandafter\global\expandafter\let
        \csname arxiv@body@\arxiv@thislabel\endcsname\BODY
      \expandafter\xdef\csname arxiv@type@\arxiv@thislabel\endcsname{#1}%
    \fi
  \else\ifcsname arxiv@partner@\arxiv@thislabel\endcsname
    \expandafter\global\expandafter\let
      \csname arxiv@body@\arxiv@thislabel\endcsname\BODY
    \expandafter\xdef\csname arxiv@type@\arxiv@thislabel\endcsname{#1}%
    \arxiv@prefetchpartner
    \ifcsname arxiv@body@\csname arxiv@partner@\arxiv@thislabel\endcsname\endcsname
      \arxiv@emitpair
        {\csname arxiv@pairfirst@\arxiv@thislabel\endcsname}%
        {\csname arxiv@pairsecond@\arxiv@thislabel\endcsname}%
    \fi
  \else
    \csname arxiv@#1\endcsname[#2]%
      \edef\arxiv@curlabel{\arxiv@thislabel}%
      \let\caption\arxiv@caption
      \BODY
    \csname arxiv@end#1\endcsname
  \fi\fi}
\newcommand{\arxivslotwidth}[2]{\@namedef{arxiv@slotw@#1}{#2}}
\newcommand{\arxiv@slot}[1]{%
  \begin{minipage}[t]{\ifcsname arxiv@slotw@#1\endcsname
      \csname arxiv@slotw@#1\endcsname\else\arxivpairslotwidth\fi}%
    \vspace{0pt}
    \arxiv@inpairtrue
    \columnwidth=\linewidth
    \edef\@captype{\csname arxiv@type@#1\endcsname}%
    \edef\arxiv@curlabel{#1}%
    \let\caption\arxiv@caption
    \csname arxiv@slotsetup@#1\endcsname
    \csname arxiv@body@#1\endcsname
  \end{minipage}}
\newcommand{\arxivslotsetup}[2]{\@namedef{arxiv@slotsetup@#1}{#2}}
\newcommand{\arxiv@emitpair}[2]{%
  \edef\arxiv@a{#1}\edef\arxiv@b{#2}%
  \arxiv@figure[t]
    \centering
    \arxiv@slot{\arxiv@a}\hfill\arxiv@slot{\arxiv@b}%
  \arxiv@endfigure
  \expandafter\global\expandafter\let\csname arxiv@done@\arxiv@a\endcsname\relax}

\newif\ifarxiv@prereading
\newcommand{\arxivprefetch}[2]{\@namedef{arxiv@prefetch@#1}{#2}}
\newcommand{\arxiv@prefetchpartner}{%
  \edef\arxiv@p{\csname arxiv@partner@\arxiv@thislabel\endcsname}%
  \ifcsname arxiv@prefetch@\arxiv@p\endcsname
    \ifcsname arxiv@body@\arxiv@p\endcsname\else
      \begingroup
        \arxiv@prereadingtrue
        \input{\csname arxiv@prefetch@\arxiv@p\endcsname}%
      \endgroup
    \fi
  \fi}

\newcommand{\arxivcaption}[2]{\@namedef{arxiv@cap@#1}{#2}}
\def\arxiv@curlabel{}
\AtBeginDocument{\LetLtxMacro{\arxiv@origcaption}{\caption}}
\newcommand{\arxiv@caption}[2][]{%
  \ifcsname arxiv@cap@\arxiv@curlabel\endcsname
    \arxiv@origcaption{\csname arxiv@cap@\arxiv@curlabel\endcsname}%
  \else
    \arxiv@origcaption{#2}%
  \fi}
\DeclareCaptionFont{noorphan}{\parfillskip=0pt plus .75\linewidth\relax}
\RenewEnviron{figure}[1][tbp]{\arxiv@float{figure}{#1}}
\RenewEnviron{table}[1][tbp]{\arxiv@float{table}{#1}}
\RenewEnviron{figure*}[1][tbp]{\arxiv@float{figure*}{#1}}
\RenewEnviron{table*}[1][tbp]{\arxiv@float{table*}{#1}}
\AtEndDocument{%
  \@for\arxiv@l:=\arxiv@pairlabels\do{%
    \ifcsname arxiv@body@\arxiv@l\endcsname
      \ifcsname arxiv@done@\csname arxiv@pairfirst@\arxiv@l\endcsname\endcsname\else
        \PackageWarningNoLine{arxiv}{Float `\arxiv@l' was held for pairing
          but its partner never appeared; it is missing from the PDF}%
      \fi
    \fi}}

\arxivpair{fig:noise}{fig:audit}
\arxivpair{fig:overfitting}{fig:tau-growth}
\arxivpair{fig:quality-selection}{tab:selection}
\arxivprefetch{tab:selection}{results/sec56/selection_table}
\arxivslotwidth{fig:quality-selection}{0.49\linewidth}
\arxivslotwidth{tab:selection}{0.47\linewidth}
\arxivslotsetup{tab:selection}{}

\arxivcaption{fig:audit}{Gain audit; each point is one model--benchmark pair. (a)~Gain of the unevolved baseline plus one rule vs.\ the reported gain; dashed line: $y{=}x$. (b)~Gain of the same artifacts under the repaired evaluation. Benchmark groups and per-pair values: Appendix~\ref{app:audit:results}. \emph{The largest reported gains are reproduced by one sentence and vanish after repair.}}

\arxivcaption{fig:quality-selection}{Headroom in the pool (a) and what selectors capture of it (b) on Terminal-Bench v2.1. (a)~Quality@$K$ (solid) vs.\ matched equal-quality null (dashed) on the 10 disjoint test tasks ($n{=}100$ skills, $\kappa_{\max}{=}100$ EB), contrasting the 2-hour budget (blue) and original timeouts (charcoal); shading: 95\% intervals. (b)~Selected-artifact gain on the 11-task cross-source selection pool (Appendix~\ref{app:selection}); Random draws an artifact, and the oracle is clipped to the selector range.}
\arxivcaption{tab:variance-text}{Variance measurement for every configuration, in pp, with the quantities and estimators defined in Appendix~\ref{app:variance-results}. Runs$\times$evals: independent evolution runs per method and test evaluations per artifact. --: not run (coverage in Table~\ref{tab:coverage}). $\dagger$: every run returned the initial harness, so this is not a method configuration.}
\arxivcaption{fig:rq1-ridge}{Distribution of the score a single run would report (Appendix~\ref{app:variance-results}) for the unevolved baseline (Base) and the four text-artifact methods (SOL: \sol{}, GEPA: \gepa{}, RB: \rbank{}, T2S: \tts{}), from $10^5$ resamples of the observed artifacts and evaluations. Dotted line: mean over all runs and evaluations. Ridges share one height; horizontal ranges differ between panels.}
\arxivcaption{tab:margins}{Survey of reported margins (Appendix~\ref{app:margins}). Columns: benchmark--model columns with an optimization competitor. Margin: median, in pp, of the proposed method's score minus its strongest competitor's. Runs: evolution runs behind each reported score, as stated in the paper; ``not stated'' marks a single reported score, which we treat as a single run.}
\arxivcaption{tab:audit-full}{Scores behind the gain audit, in pp, averaged over evaluation repetitions. Released: the evaluation released with the method; baseline + one rule: scored under the released evaluation; repaired: after the repair in Table~\ref{tab:audit-repairs}. Bold: 95\% confidence interval excludes zero. --: no repair, no one-rule result, or not run (Appendix~\ref{app:audit}). On LiveMathematicianBench, repaired scores use the 60 of the 106 test questions without the leaked option.}
\makeatother

\newif\iflanshaocheng
\lanshaochengfalse
\iflanshaocheng
  \newcommand{\lyx}[1]{\textcolor{blue}{[LI: #1]}}
  \newcommand{\kan}[1]{\textcolor{orange}{[REN: #1]}}
  \newcommand{\zz}[1]{\textcolor{green}{[ZZ: #1]}}
  \newcommand{\lsc}[1]{\textcolor{purple}{[Shaocheng: #1]}}
  \newcommand{\todo}[1]{\textcolor{red}{[TODO: #1]}}
  \newcommand{\ph}[1]{\textcolor{red}{[#1]}}
  \newcommand{\la}[1]{\textcolor{brown}{[LA: #1]}}
  \newcommand{\zyh}[1]{\textcolor{red}{[ZYH: #1]}}
\else
  \newcommand{\lyx}[1]{}
  \newcommand{\kan}[1]{}
  \newcommand{\zz}[1]{}
  \newcommand{\lsc}[1]{}
  \newcommand{\todo}[1]{}
  \newcommand{\ph}[1]{}
  \newcommand{\la}[1]{}
  \newcommand{\zyh}[1]{}
\fi

\newcommand{\sig}{\sigma}
\newcommand{\taug}{\tau}
\newcommand{\sol}{SkillOpt-Lite}
\newcommand{\gepa}{GEPA}
\newcommand{\rbank}{ReasoningBank}
\newcommand{\tts}{Trace2Skill}
\newtcolorbox{hypothesisbox}{colback=gray!6,colframe=gray!55,arc=1.5mm,boxrule=0.6pt,left=2mm,right=2mm,top=1.5mm,bottom=1.5mm}

\newif\ifresultdraft
\resultdraftfalse

%% file: math_commands.tex

\usepackage{amsmath,amsfonts,bm}









\def\eqref#1{equation~\ref{#1}}









\def\1{\bm{1}}










\DeclareMathAlphabet{\mathsfit}{\encodingdefault}{\sfdefault}{m}{sl}
\SetMathAlphabet{\mathsfit}{bold}{\encodingdefault}{\sfdefault}{bx}{n}













%% file: metadata.tex
\title{The Resolution Hypothesis:\\Agent Evolution's Key Is Identifying Improvements}

\author[1,2,3,*]{Yuanhao Zeng}
\author[1,2,*]{Shaocheng Lan}
\author[1,2,*]{Ao Lu}
\author[2]{Kuang Yan}
\author[2]{Heng Lin}
\author[2]{Xinang Xu}
\author[4]{Yuan Luo}
\author[1]{Yueting Wang}
\author[3]{Yexin Li}
\author[2]{Ruo Shan}
\author[1,\dagger]{Kan Ren}

\affiliation[1]{ShanghaiTech University}
\affiliation[2]{ByteDance, China}
\affiliation[3]{BIGAI}
\affiliation[4]{SUSTech}

\contribution[*]{Equal contribution}
\contribution[\dagger]{Corresponding author}
\correspondence{\email{zengyh2025@shanghaitech.edu.cn}, \email{renkan@shanghaitech.edu.cn}}
\checkdata[Project Page]{\url{https://lineshogan.github.io/resolution-hypothesis/}}

%% file: sections/00_abstract.tex
\begin{abstract}
Agent evolution follows a simple loop: a proposer writes candidate artifacts (prompts, skills, memories) and selection accepts those that score higher.
This loop has shown promise, yet practitioners face a puzzle: methods trade wins across papers, and evolution often fails to bootstrap on certain tasks.
We trace this puzzle to the \emph{Resolution Hypothesis}: agent evolution stalls not because methods fail to propose better artifacts, but because noisy evaluations cannot reliably identify them.
Across seven evolution methods and six models (over 10M LLM calls), we find support for the hypothesis at both levels where evaluation scores are used.
At the candidate level (choosing artifacts), helpful candidate artifacts exist, but selection is no better than random at telling them from harmful ones, so errors accumulate across rounds.
At the method level (ranking methods), typical margins of a few percentage points are smaller than the difference between two independent runs of the same method, so published rankings often rest on luck.
Together, these findings indicate that agent evolution is often bottlenecked by reliably identifying improvements. This calls for a shift in how we develop and evaluate self-improving agents, from designing ever more candidate improvement systems to making true improvements detectable and accumulable.
\end{abstract}

%% file: content.tex
\input{sections/01_introduction_new}
\input{sections/02_related_work_new}
\input{sections/03_measurement_model}
\input{sections/04_evaluation_crisis}

\input{sections/05a_audit}
\input{sections/05_proposals_exist}
\input{sections/06_selection_fails}
\input{sections/07_noise_accumulates}
\input{sections/08_ways_forward}
\input{sections/10_conclusion}
\input{sections/09_boundary_limitations}
\input{sections/11_use_of_ai}

\FloatBarrier
\bibliographystyle{plainnat}
\bibliography{\arxivpath custom}

\clearpage
\appendix
\input{sections/app_related_work}

\section{Variance Decomposition}
\label{app:decomposition}
\input{sections/app_variance_decomposition}
\input{sections/app_estimators}
\input{sections/app_setup}

\FloatBarrier
\section{Variance Measurement}
\label{app:variance}
\input{sections/app_variance_results}

\input{sections/app_power}
\input{sections/12a_appendix_margins}
\input{sections/12_appendix_audit}

\FloatBarrier
\section{Candidate Pool and Selection}
\label{app:candidate-pool}
\input{sections/app_pool}
\input{sections/app_qualityatk}
\input{sections/app_selection}
\input{sections/12b_appendix_grouped_curves}
\FloatBarrier
\input{sections/app_accumulation}
\input{sections/app_wayout}

\FloatBarrier
\section{Case Study}
\input{sections/app_shared_artifacts}

%% file: sections/01_introduction_new.tex
\section{Introduction}

\begin{figure}[t]
\centering
\includegraphics[width=\columnwidth]{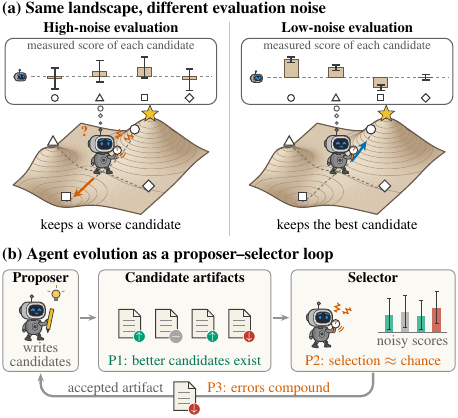}
\caption{(a)~Four candidate moves on the same landscape, scored under high and low evaluation noise (thought bubble: score relative to the current artifact). Under high noise, a worse candidate artifact scores highest. (b)~Agent evolution as a loop: a proposer writes candidate artifacts, and a selector accepts the artifacts with the highest noisy score. P1--P3 are the predictions we test in Section~\ref{sec:inside}.}
\label{fig:teaser}
\end{figure}

Agent evolution improves frozen LLM agents by refining persistent external artifacts such as prompts, skills, memories, and entire agent harnesses.
It typically follows a loop as shown in Figure~\ref{fig:teaser}b: a proposer writes candidate artifacts, the resulting agent is evaluated on a task set, and selection accepts the candidate artifacts that score higher.
Evolution methods of this family have shown promise across benchmarks and models~\citep{skillopt2026,ouyang2026reasoningbank,trace2skill2026}, making the prospect of systems that continuously improve their own workflows increasingly plausible.

Yet the picture is far from settled.
Studies often reach conflicting conclusions: the same methods flip between effective and ineffective across benchmarks, and their rankings trade places across papers.
On Terminal-Bench~\citep{terminalbench} and other coding-agent benchmarks, evolution methods have repeatedly been reported to stall~\citep{shang2026selfevolutionbackfires,li2026skillsbench,dong2026agentskillsharmfulempirical}, while on other benchmarks each new method claims to outperform its predecessors~\citep{ace,lin2026ahe,trace2skill2026,skilloptlite}.
Whether evolution helps depends on the setting in ways the community cannot yet predict.

This instability erodes confidence in this direction in both academia and industry, and rewards whichever method happened to be luckier at evaluation time.
Yet unreliable evaluation affects more than the reported rankings of evolution methods.
At the candidate level, the same kind of evaluation decides which artifacts to carry forward to the next round.
As a result, the evolution methods themselves become unreliable.

We argue that this evaluation rests on an untested premise that its scores can identify the improvements evolution brings. 
We find that the premise fails at both levels where the scores are used. 
At the method level, the margins that separate methods in papers are smaller than the difference between two independent runs of the same method: 
on Terminal-Bench, current common practice (Appendix~\ref{app:margins}) can only identify the truly best of four text-artifact methods 33\% of the time, barely above a random guess (Section~\ref{sec:crisis}). 
At the candidate level, the selector picks the better of two candidate artifacts only about half the time, no better than a coin flip (Section~\ref{sec:selection}). 
Because each round accepts the artifact that happened to score highest, the scores go up but the artifacts do not get better (Section~\ref{sec:accumulation}).

Both failures trace to a single constraint, which we state as a hypothesis.

\begin{hypothesisbox}
\textbf{The Resolution Hypothesis.} \emph{The bottleneck of agent evolution is identifying which artifacts bring improvement, not proposing them.}
\end{hypothesisbox}

\noindent Why is identifying improvements so hard? Because model weights are frozen, an artifact helps only by changing the model's behavior at inference time, while LLM trajectories also vary each time the agent attempts a task.
As a result, separating a true behavioral change from evaluation noise is difficult. We define the \emph{resolution} of an evaluation pipeline as the smallest true improvement it can reliably distinguish from noise.
Once the true improvement falls below this resolution, candidate selection degrades to chance, so little improvement accumulates during evolution (Figure~\ref{fig:teaser}a).
Proposal determines how large the improvements on offer are; resolution determines how small an improvement can be and still be reliably identified.

We test this hypothesis at a much larger scale than current practice. 
Whereas prior work typically relies on a single evolution run (Appendix~\ref{app:margins}), we run each method 8 times per model and benchmark and evaluate each resulting artifact up to 16 times, totaling 1,030 independent evolution runs and over 2.1 million test trajectories.
In total, the study covers seven agent evolution methods over six models and four agent benchmarks, plus nine further benchmarks on which we re-examine previously reported gains; to our knowledge, it is the largest reproducibility study of agent evolution to date. Our findings are:
\begin{itemize}[leftmargin=3mm]
\item \textbf{Method level (Section~\ref{sec:crisis}).}
Typical method margins are smaller than what a single run can identify.
The large historically reported gains are the clearest instance of our hypothesis: on those benchmarks, evolution succeeds by exploiting evaluation shortcuts that lift nearly every task at once, producing gains large enough to exceed the resolution of the evaluation, and stalls once the evaluation is repaired (Section~\ref{sec:audit}).
\item \textbf{Candidate level (Section~\ref{sec:inside}).}
We derive three predictions from the hypothesis. All three are supported by our experiments on Terminal-Bench:
\begin{itemize}
\item[\textbf{P1}] Candidate pools contain artifacts that genuinely improve task performance: proposal is not the bottleneck (Section~\ref{sec:proposals}).
\item[\textbf{P2}] Common selection mechanisms---whether based on execution scores or LLM judgments---perform near chance at identifying the better of two candidate artifacts (Section~\ref{sec:selection}).
\item[\textbf{P3}] Per-round selection errors compound across rounds, producing the large differences between independent runs of the same method that flip rankings (Section~\ref{sec:accumulation}).
\end{itemize}
\end{itemize}

Finally, we explore practical ways to address low evaluation resolution at both the candidate and method levels, from lower-variance behavioral checks to separate evaluations of proposers and selectors (Section~\ref{sec:forward}).

In conclusion, these findings suggest a different perspective on agent evolution. The proposer determines what improvements are available, but evaluation resolution determines which improvements can be retained and accumulated. When this resolution is insufficient, true improvements are lost in noise, while large, easily detectable changes can dominate the evolution process even when they exploit evaluation shortcuts. 
Progress requires either enlarging the improvements being proposed or making smaller improvements reliably detectable. 
More broadly, developing higher-resolution feedback may be as important as developing more powerful evolution algorithms.

All reproducible materials and trajectory data will be released upon the acceptance of this paper.

%% file: sections/02_related_work_new.tex
\section{Related Work}
\label{sec:related}

Agent evolution methods share a common propose-evaluate-select loop but vary in what they evolve: prompts~\citep{agrawal2025gepareflectivepromptevolution}, skills~\citep{trace2skill2026,skillopt2026}, memories~\citep{ace,ouyang2026reasoningbank}, or executable code~\citep{lin2026ahe,dgm}.
Several analyses document the fragility of this paradigm, e.g., sensitivity to run-to-run variance and task order~\citep{fragility}, sparse feedback-dependent gains~\citep{liu2026rethinkingselfevolvingagentskills}, and skill-induced failures or contamination from uncontrolled accumulation~\citep{li2026skillsbench,dong2026agentskillsharmfulempirical,shang2026selfevolutionbackfires}.
These findings point to noisy selection as a shared failure mode, connecting to best-arm identification~\citep{bestarm}, uncertainty-aware RL evaluation~\citep{agarwal2021deep}, and the false-commit analysis of PACE~\citep{shawn2026pace}.
Our work isolates the mechanism: we separate evaluation noise from run-to-run variance and use controlled pools to test proposal quality and selection reliability independently.
Appendix~\ref{app:related} gives a full discussion.

%% file: sections/03_measurement_model.tex
\section{Variance Decomposition and Experimental Setup}
\label{sec:setup}

To determine whether benchmark scores can identify improvements, we first need a formal model that separates the stochasticity of the evolution algorithm from the noise of the evaluation instrument.
This section defines the variance decomposition and outlines our experiments.

\subsection{Two Sources of Score Variance}
\label{sec:setup:model}

Consider an evolution method evaluated on a benchmark.
Let artifact $A$ denote the persistent external state produced by the method (a prompt, a skill library, a memory playbook, or an agent harness).
Evaluating artifact $A$ once yields an observed score $\hat{Y}(A)$.
We define the true quality of artifact $A$ as $Y(A) = \mathbb{E}[\hat{Y}\mid A]$, its expected score over repeated evaluations.
A single observed score decomposes as $\hat{Y}(A) = Y(A) + \epsilon$, where $\epsilon$ is zero-mean evaluation noise ($\mathbb{E}[\epsilon\mid A] = 0$).
Across independent evolution runs, the produced artifacts vary; let $\tau^2 = \mathrm{Var}(Y(A))$ denote this run-to-run variance in true quality.
By the law of total variance, the variance of an observed score decomposes as:
\begin{equation}
\mathrm{Var}(\hat{Y}) = \underbrace{\mathbb{E}_{A}\!\left[\mathrm{Var}(\hat{Y}\mid A)\right]}_{\sigma^2} + \underbrace{\mathrm{Var}_{A}\!\left(\mathbb{E}[\hat{Y}\mid A]\right)}_{\tau^2}.
\label{eq:decomposition}
\end{equation}
The first term $\sigma^2$ averages over evaluation noise within each artifact; the second $\tau^2$ measures how much artifact quality itself varies across runs.
Standard practice often reports a single run evaluated once (Table~\ref{tab:margins}), conflating both sources. To identify true improvements, each source must be bounded separately.
Crucially, the resolution depends on the question. Identifying the gain of a single fixed artifact over a reference is limited by $\sigma^2$ alone, while identifying a difference between two methods is limited by $(\tau^2+\sigma^2)$, because each run of a method draws a new artifact.

\subsection{Experiments}
\label{sec:setup:grid}

We study seven evolution methods in two families. Four \emph{text-artifact methods} evolve text given to the agent: prompts (\gepa{}~\citep{agrawal2025gepareflectivepromptevolution}), skills (\sol{}~\citep{skilloptlite}, \tts{}~\citep{trace2skill2026}), and memory playbooks (\rbank{}~\citep{ouyang2026reasoningbank}). Three \emph{harness methods} evolve the agent's scaffolding (HarnessLens~\citep{xu2026verifysmarterevolvefurther}, Meta-Harness~\citep{lee2026metaharnessendtoendoptimizationmodel}, AHE~\citep{lin2026ahe}).
We use six agent models: GPT-5.4 nano~\citep{openai2026gpt54nano}, GPT-5.6 Luna and Terra~\citep{openai2026gpt56}, DeepSeek-V4-Flash~\citep{deepseekai2026deepseekv4}, Kimi-K3~\citep{kimiteam2026kimik3}, and GLM-5.2~\citep{glmteam2026glm5,zai2026glm52}.
Model coverage follows cost: the cheaper experiments use all six models and the more expensive ones a subset.
The experiments fall into four groups:
\begin{itemize}[leftmargin=3mm]
\item \textbf{Variance measurement} (Section~\ref{sec:crisis:repetitions}). All seven methods on ALFWorld~\citep{alfworld}, SpreadsheetBench~\citep{spreadsheetbench}, OfficeQA~\citep{officeqa}, and Terminal-Bench v2.1~\citep{terminalbench}, each evolved and evaluated repeatedly to estimate $\tau$ and $\sigma$.
\item \textbf{Gain audit} (Section~\ref{sec:audit}). The largest reported gains on twelve benchmarks, the first three above and nine from the methods' original papers, re-examined under the released and repaired evaluations; on four of them, we also re-evolve the methods under the repaired evaluations.
\item \textbf{Candidate pools} (Sections~\ref{sec:proposals} and~\ref{sec:selection}). Without loss of generality, we use text-based skills to test proposal quality and selection reliability in isolation, constructing two pools of $n{=}100$ skills tailored to each question (Appendix~\ref{app:pool}): a \textbf{Proposal Pool} (Section~\ref{sec:proposals}) generated on 10 Terminal-Bench v2.1 training tasks and evaluated on 10 disjoint test tasks (16 rollouts per pair) to measure out-of-sample proposal gains, and a \textbf{Selection Pool} (Section~\ref{sec:selection}) combining candidates from Terminal-Bench and DeepSWE trajectories across 11 difficulty-stratified Terminal-Bench tasks (8 to 15 rollouts per pair) to benchmark selection rules and trajectory judges.
\item \textbf{Trajectory re-evaluation} (Section~\ref{sec:accumulation}). Every intermediate artifact of eight \sol{} and eight \gepa{} evolution runs on Terminal-Bench, re-evaluated eight times per task.
\end{itemize}
Appendix~\ref{app:setup} gives the setup shared by all experiments and the coverage of each group (Table~\ref{tab:coverage}); each group's protocol and full results are in its own appendix, listed in the table.
In total, the experiments consumed over 10M LLM calls and 400K CPU core hours.

%% file: sections/04_evaluation_crisis.tex
\begin{figure}[t]
\centering
\includegraphics[width=\columnwidth]{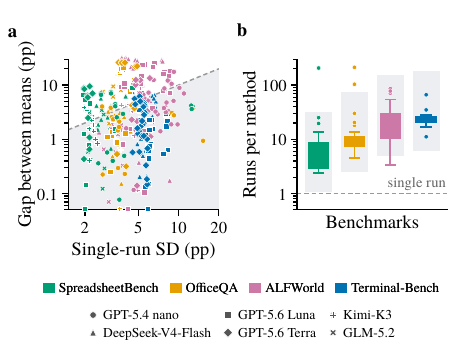}
\caption{(a)~Gap between all-run means of same-benchmark configuration pairs vs.\ the standard deviation of a single-run comparison; shaded: gap below one standard deviation. (b)~Runs per method to identify the median literature margin (3.4 pp) at 80\% power; gray bar: interquartile range 1.2--6.6 pp. Per-method values: Table~\ref{tab:variance-text}. \emph{Typical method differences are within single-run resolution.}}
\label{fig:noise}
\end{figure}

\section{Resolution at the Method Level}
\label{sec:crisis}

With both variance components measured, we can test whether standard evaluation practice can identify the method differences these papers report.

\subsection{Single Reports Cannot Rank Methods}
\label{sec:crisis:repetitions}

For each method, model, and benchmark, we run evolution 8 times independently (16 times for the text-artifact methods on GPT-5.4 nano) and evaluate each resulting artifact 16 times on the held-out test split (8 times on Terminal-Bench and for harness methods).
The unevolved baseline is evaluated equally often.
We call a method or the unevolved baseline on a given model and benchmark a \emph{configuration}; in total, these experiments comprise 150 configurations, 1{,}030 independent evolution runs, and over 2.1 million test trajectories.
We estimate $\sigma$ from how much repeated evaluations of the same artifact differ, and $\tau$ from how much the mean scores of the artifacts from different runs differ (Appendix~\ref{app:estimators}).

\paragraph{Most method differences are below one single-run standard deviation.}
The standard deviation of the evaluation noise, $\sigma$, ranges from 1.0 to 3.7 percentage points (pp) on SpreadsheetBench, OfficeQA, and ALFWorld, and reaches 4.5 pp on Terminal-Bench, while run-to-run standard deviation $\tau$ ranges from 0 to 12.3 pp.
Together, a method evolved once and evaluated once has a typical score standard deviation of 1.7 to 4.1 pp, the median over the configurations of each benchmark (values for every configuration: Table~\ref{tab:variance-text}).
Most method differences are smaller than this: for over 60\% of pairs of configurations on the same model and benchmark, the gap between their all-run means falls below one standard deviation of a single-run comparison (Figure~\ref{fig:noise}a).
The result holds method by method: for every method, 53\% to 79\% of its comparisons fall below one single-run standard deviation (Appendix~\ref{app:variance-results}).

\paragraph{Reported margins fall in the same range.}
Across 171 benchmark--model results in the main tables of nine agent-evolution papers, the proposed method leads its strongest competitor by a median of 3.4 pp, with an interquartile range of 1.2 to 6.6 pp.
None of these results reports a variance estimate, and only 6 average over more than one evolution run. Appendix~\ref{app:margins} lists the surveyed results.

\paragraph{Identifying such margins takes many runs.}
Identifying a true margin $\Delta$ at 80\% power with $\alpha = 0.05$ requires
\begin{equation}
N \ge 2 \cdot \left(\frac{z_{1-\alpha/2} + z_{1-\beta}}{\Delta}\right)^2 \cdot (\tau^2 + \sigma^2)
\label{eq:power}
\end{equation}
independent runs per method.
At the median reported margin of 3.4 pp, this means about 5 runs on SpreadsheetBench, 10 on OfficeQA, 19 on ALFWorld, and 23 on Terminal-Bench (181 on Terminal-Bench at the lower-quartile margin) (Figure~\ref{fig:noise}b).
A single run cannot rank methods whose true difference is this small: a results table with one run per method picks the best of the four text-artifact methods only 51\% to 73\% of the time on SpreadsheetBench, OfficeQA, and ALFWorld, and only 33\% on Terminal-Bench, where evaluation noise is largest; a random guess gets 25\%.
Appendix~\ref{app:power} gives the full distribution and the ranking simulation.

%% file: sections/05a_audit.tex
\subsection{Large Reported Gains Come from Evaluation Shortcuts}
\label{sec:audit}

Section~\ref{sec:crisis:repetitions} shows that typical method differences are too small for current evaluation to identify.
Some benchmarks, however, show gains of 10 to 40 pp, far above the resolution of a single run, both in the original papers and in our own runs, for example on SpreadsheetBench and LiveMathematicianBench (Appendix~\ref{app:audit:design}). This suggests that evolution can find large improvements after all.
The Resolution Hypothesis predicts exactly this pattern: evolution can accumulate an improvement only if the true improvement exceeds the resolution.
An evaluation shortcut can produce such an improvement: a single behavior that the evaluation rewards on most tasks of a benchmark at once yields a gain large enough for even a noisy evaluation to identify.
We test this prediction by auditing the largest reported gains on twelve benchmarks with three models, using two tests (design in Appendix~\ref{app:audit}).

\begin{figure}[t]
\centering
\includegraphics[width=\columnwidth]{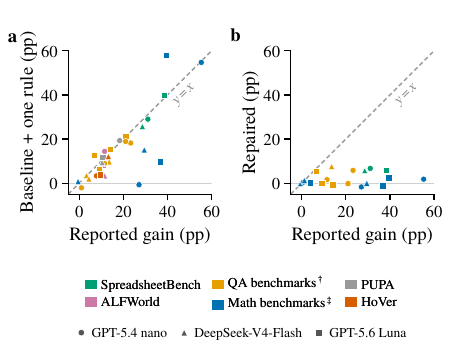}
\caption{(a)~Unevolved baseline plus one rule. (b)~Same artifacts under the repaired evaluation (per-pair values: Appendix~\ref{app:audit:results}). \emph{The largest reported gains are reproduced by one sentence and vanish after repair.}}
\label{fig:audit}
\end{figure}

\paragraph{One sentence reproduces the gain.}
Evolved artifacts contain many instructions, so a large gain could come from many improvements or from a single one.
To find out, we audit, instruction by instruction, the artifacts released with \gepa{} and \sol{} and those we evolved ourselves (an LLM agent flags candidate instructions, and we check each by hand), and append to the unevolved baseline only the one rule that appears most responsible for the gain.
This single sentence nearly matches what full evolution achieved (Figure~\ref{fig:audit}a): across the 27 model--benchmark pairs with a reported gain above 3 pp, it recovers a median of 90\% of the gain.
The rule typically takes an evaluation shortcut: it adapts to a grader that credits only one answer format, or to a version of the benchmark, released with a method, that departs from the official evaluation (Table~\ref{tab:audit-repairs}).
Appendix~\ref{app:audit:results} discusses the exceptions.

\paragraph{Repairing the evaluation removes it.}
We then repair each evaluation that admits a shortcut, restoring the official evaluation or grading answers by meaning rather than surface form, and re-score the same artifacts.
The median gain falls from 17.6 to 0.8 pp (Figure~\ref{fig:audit}b), and on several benchmarks the repaired baseline leaves little headroom at all.
Re-running the text-artifact methods under the repaired evaluations does not bring the gain back (Appendix~\ref{app:audit:reevolve}).

This controlled contrast supports our argument: evolution succeeds when an artifact delivers an improvement large enough to exceed the resolution, and fails when only small, task-specific improvements remain.
Two questions follow: do such small improvements exist at all, and if so, why does selection fail to accept them?

%% file: sections/05_proposals_exist.tex
\section{Resolution at the Candidate Level}
\label{sec:inside}

Progress within an evolution run depends on two conditions: proposals must offer true improvements, and selection must identify them.
We examine both conditions on frozen candidate pools (Sections~\ref{sec:proposals} and~\ref{sec:selection}), then follow how selection errors behave across rounds (Section~\ref{sec:accumulation}).

\subsection{Helpful Artifacts Exist}
\label{sec:proposals}

\textbf{Testing P1.}
To observe what a proposer produces before selection filters candidates out, we construct the \textbf{Proposal Pool}: $n{=}100$ candidate skills generated offline by a \sol{}-based proposer from trajectories on 10 Terminal-Bench v2.1 training tasks, and evaluated across 10 disjoint test tasks with 16 independent rollouts per skill--task pair (16,000 rollouts total; protocol in Appendix~\ref{app:pool}).

\paragraph{More candidates yield real gains.}
\label{sec:proposals:results}

\begin{figure}[t]
\centering
\includegraphics[width=\columnwidth]{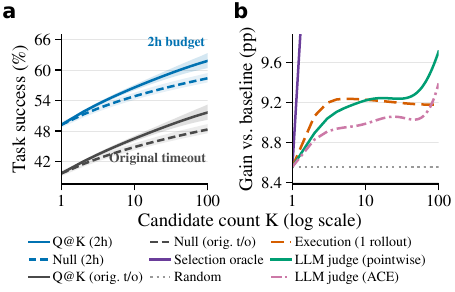}
\caption{Candidate quality in the Proposal Pool (a) and what selectors recover on the Selection Pool (b). (a)~Quality@$K$ (solid) vs.\ matched equal-quality null (dashed) across the 10 disjoint Terminal-Bench v2.1 test tasks of the Proposal Pool ($n{=}100$ skills, 16 rollouts each), contrasting the 2-hour budget (blue) against the original per-task timeout (charcoal) under the $\kappa_{\max}{=}100$ empirical-Bayes estimator; shading shows pointwise 95\% posterior credible intervals (observed) and 95\% null simulation ranges. (b)~Selected-artifact gain over baseline vs.\ candidate budget $K$ across the 11 tasks of the Selection Pool (Appendix~\ref{app:pool}; held-out quality reference in Appendix~\ref{app:selection}); random draws an artifact rather than the baseline, and the oracle is clipped to the selector range.}
\label{fig:quality-selection}
\end{figure}

If a proposer writes $K$ candidate skills and we keep the strongest one for each task, does task performance genuinely improve as $K$ grows?
The solid curves in Figure~\ref{fig:quality-selection}a track this quantity, \emph{Quality@$K$}, defined as the expected success rate of the best among $K$ randomly drawn skills.
Under a 2-hour execution budget (blue solid curve), which allows multi-step solutions to run to completion, Quality@$K$ climbs steadily from $49.2\%$ at $K{=}1$ to $61.8\%$ at $K{=}100$.
Under Terminal-Bench's shorter original timeouts (charcoal solid curve), overall success starts lower ($39.7\%$ at $K{=}1$) because slow trajectories are cut short, yet Quality@$K$ still rises to $51.6\%$ at $K{=}100$.

Because taking the maximum over $K$ noisy evaluations would curve upward even if all 100 skills were identical, each solid curve in Figure~\ref{fig:quality-selection}a is paired with a dashed \emph{equal-quality null}---the curve produced when all 100 candidates share the same true success rate and pass through the same empirical-Bayes estimator (Appendix~\ref{app:qualityatk}).
The vertical gap between the solid and dashed curves isolates genuine skill differences from sampling luck.
In both timeout regimes, the observed curve pulls clearly above its dashed null by $K{=}5$ and maintains a $+3.3$ to $+3.4$ pp lead at $K{=}100$ (with intermediate timeout cutoffs in Appendix~\ref{app:pool}).
Generating more candidates thus yields genuinely stronger skills, with gains comparable to the post-repair method margins in Section~\ref{sec:audit}.

\paragraph{Shareable artifacts exist.}
While Quality@$K$ allows each task to pick its own best skill, a single skill can also improve performance across all 10 test tasks at once.
Averaged over all 160 test rollouts per candidate, individual skills in the Proposal Pool span more than 15 percentage points: under the 2-hour budget, the strongest shared skill (\texttt{artifact-063}) reaches $57.5\%$ compared with the $49.2\%$ pool mean ($+8.3$ pp) and $41.9\%$ for the weakest skill, with a similar $+8.4$ pp lead under the original timeout ($48.1\%$ vs.\ $39.7\%$).
Appendix~\ref{app:shared-artifacts} inspects representative shared and task-specific skills from our candidate pools to show what these improvements look like in practice.

Pools of this kind are inconsistent with the claim that LLMs cannot propose effective artifacts.
Because proposal alone does not explain why evolution stalls, we next isolate selection in Section~\ref{sec:selection}.

%% file: sections/06_selection_fails.tex
\subsection{Selection Cannot Identify Them}
\label{sec:selection}
\input{results/sec56/numbers.tex}

\textbf{Testing P2.}
Because candidate pools contain helpful skills (Section~\ref{sec:proposals}) yet evolution fails to accumulate them, we next isolate the selection step.
We construct the \textbf{Selection Pool}: $n{=}100$ skills generated from trajectories on Terminal-Bench and DeepSWE, evaluated across 11 Terminal-Bench tasks spanning five difficulty levels (Appendix~\ref{app:pool}).
Each selector observes only the first rollout (two rollouts for native \sol{} and \gepa{} selection), while true skill quality is measured on the remaining held-out rollouts (Appendix~\ref{app:selection}).
We test whether standard selectors choose better skills more often than chance.

\paragraph{Under realistic budgets, selection operates near chance.}
\label{sec:selection:budget}

Greedy accept/reject on execution scores, if effective, would accept improvements and reject regressions with high reliability.
Table~\ref{tab:selection} and the rank correlation below show the opposite on all three metrics.
In head-to-head comparisons, execution-based selection prefers the better artifact only 49.7\% of the time, statistically indistinguishable from the 50\% achieved by random selection.
Regressions make up 30.5\% of the pool but 62.1\% of accepted candidates, so selection does not filter out regressions; the wide interval leaves open whether it is worse than random acceptance.
The rank correlation between a single execution score and posterior artifact quality is \ExecRank, essentially zero.
The native parent-selection operators of \sol{} and \gepa{}, which observe two rollouts per artifact rather than one, achieve 48.6\% pairwise accuracy.
Every execution-based selector in this study is statistically indistinguishable from random choice.

\paragraph{No selector benefits from more candidates.}
Figure~\ref{fig:quality-selection}b plots the gain over baseline of the selected skill as the candidate budget $K$ grows from 1 to 100, averaged across the 11 tasks (per-task curves in Appendix~\ref{app:grouped-curves}).
While the \emph{Selection oracle} (purple curve at the top) climbs steadily with $K$ as better skills enter the sample, every practical selector remains nearly flat around the random-selection baseline (gray dashed line).
Even after examining all $100$ candidates, no selector recovers the quality that an oracle already reaches within $K < 5$ candidates.

\paragraph{Model-based judges do not restore identification.}
\label{sec:selection:judges}

Model-based judges are a natural candidate for restoring identification, because they read the full trajectory rather than a single terminal outcome~\citep{zheng2023judging,ace}. So one might hope they could tell helpful artifacts from harmful ones.
Table~\ref{tab:selection} shows that this additional evidence does not help.
No judge exhibits reliable pairwise accuracy.
This may be because the judge has limited ability, while its reading noise is independent of evaluation noise. The two therefore accumulate rather than cancel out, introducing additional noise into the evaluation process.

\input{results/sec56/selection_table.tex}

The consistency of this failure across such different selectors supports P2: under realistic budgets, standard selection mechanisms cannot identify helpful artifacts. Appendix~\ref{app:selection:higher} examines whether repeated validation helps.

%% file: results/sec56/numbers.tex
\newcommand{\ExecImprove}{70.064}
\newcommand{\RandomImprove}{69.675}
\newcommand{\ExecRegress}{29.936}
\newcommand{\ExecRank}{0.054}

%% file: results/sec56/selection_table.tex
\begin{table}[t]
\centering
\footnotesize
\setlength{\tabcolsep}{2pt}
\begin{tabularx}{\columnwidth}{@{}Xcc@{}}
\toprule
Selection & Pairwise acc.\,(\%) & False accept.\,(\%) \\
\midrule
Random & 50.0\,$\pm$\,0.0 & 30.5\,$\pm$\,14.0 \\
Execution (1 rollout) & 49.7\,$\pm$\,1.9 & 62.1\,$\pm$\,42.9 \\
\sol{}\textsuperscript{\dag} & 48.6\,$\pm$\,2.6 & 59.4\,$\pm$\,46.2 \\
\gepa{}\textsuperscript{\dag} & 48.6\,$\pm$\,2.6 & 31.6\,$\pm$\,14.5 \\
LLM judge (pointwise) & 45.4\,$\pm$\,4.5 & 39.9\,$\pm$\,17.5 \\
LLM judge (pairwise) & 49.0\,$\pm$\,2.2 & 57.6\,$\pm$\,23.3 \\
LLM judge (ACE) & 46.9\,$\pm$\,3.5 & 39.0\,$\pm$\,15.6 \\
\bottomrule
\end{tabularx}
\caption{Selection mechanisms evaluated at $K{=}n$ on the fixed candidate pool.
\textbf{Pairwise acc.}: probability that the selector picks the higher-quality artifact in a head-to-head comparison; 50\% is chance.
\textbf{False accept.}: among artifacts accepted as baseline replacements, the fraction that are actually worse.
Values are posterior means $\pm$ half-width of 95\% credible intervals.
\textsuperscript{\dag}Uses two visible rollouts per artifact; all others use one.
\emph{Neither execution-based selection nor LLM judges can reliably identify helpful artifacts among regressions.}}
\label{tab:selection}
\end{table}

%% file: sections/07_noise_accumulates.tex
\subsection{Noise Accumulates Across Rounds}
\label{sec:accumulation}

\textbf{Testing P3.}
Section~\ref{sec:selection} showed that individual selection steps discriminate at chance level.
An evolution run applies such a step in every round, so the relevant quantity is the cumulative effect of per-round errors over rounds.

\paragraph{In-loop gains reflect noise exploitation.}
\label{sec:accumulation:drift}

\begin{figure}[t]
\centering
\includegraphics[width=\columnwidth]{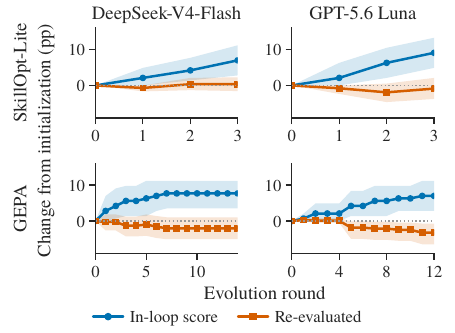}
\caption{In-loop vs.\ independently re-evaluated validation gains across rounds (relative to initialization; shading: 95\% bootstrap intervals over eight runs). \emph{In-loop gains reflect noise exploitation, not true improvement.}}
\label{fig:overfitting}
\end{figure}

Near-chance selection implies that most of the gain a run records in its own validation scores, its in-loop gain, should disappear under careful re-evaluation.
This is what the re-evaluation shows.
We independently re-evaluate every artifact produced by eight evolution runs of \sol{} and eight of \gepa{} on the same Terminal-Bench validation set, with eight rollouts per artifact--task pair where evolution used one (protocol in Appendix~\ref{app:accumulation}).
In-loop gains grow to 6.9 to 9.0 pp over the course of evolution, while re-evaluated gains remain between $+0.4$ and $-3.2$ pp (Figure~\ref{fig:overfitting}).
The mechanism follows from the acceptance rule, which compares a candidate's noisy score against the current artifact's and therefore accepts lucky draws from the upper tail of the noise distribution round after round~\citep{cawley2010,shawn2026pace}.
In-loop scores improve under this process; agent capability does not.

\paragraph{Run-to-run variance grows over rounds.}
\label{sec:accumulation:tau}

Per-round selection errors also drive independent runs apart.
All runs begin from the same artifact, so $\tau^2_0 = 0$, and each round of noisy selection is one more step of an unguided random walk over artifact space.
By the final round, the corrected run-to-run variance $\hat\tau^2_t$ reaches 4.2 to 24.5 squared pp across method--model groups (Figure~\ref{fig:tau-growth}; estimation details in Appendix~\ref{app:accumulation}).
Independent runs of one method now differ by as much as the margins that separate methods in the literature, which is the instability documented in Section~\ref{sec:crisis}.
These results support P3: per-round selection errors compound across rounds and produce the run-to-run variance behind the ranking instability between methods.

\begin{figure}[t]
\centering
\includegraphics[width=\columnwidth]{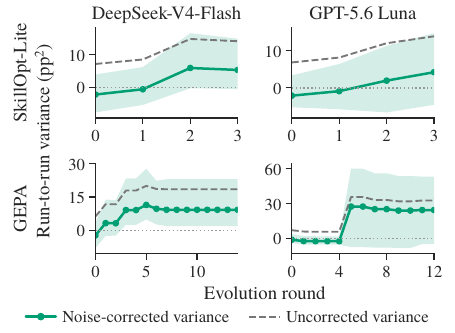}
\caption{Run-to-run variance $\hat\tau^2_t$ over rounds ($\tau^2_0 = 0$; dashed: raw; solid: noise-corrected; shading: 95\% bootstrap intervals over eight runs). \emph{Run-to-run variance grows over rounds.}}
\label{fig:tau-growth}
\end{figure}

In summary, the variance that prevents a single selection step from identifying improvement is the same variance that destabilizes rankings between methods, compounded over rounds.

%% file: sections/08_ways_forward.tex
\section{Towards Better Evaluation Resolution}
\label{sec:forward}

Because low evaluation resolution limits both candidate selection (Section~\ref{sec:inside}) and method comparison (Section~\ref{sec:crisis}), we examine two practical ways to detect improvements that terminal pass/fail scores miss: finer-grained behavioral checks at the candidate level, and separate evaluations of proposers and selectors at the method level.

\paragraph{Candidate level: finer-grained feedback.}
Terminal pass/fail rewards may obscure improvements in specific agent behaviors.
A promising direction is to supplement them with programmatic checks of intermediate behavior, such as unmasked failure diagnostics on verification commands or whether the agent runs a test after editing a file.
Such checks do not fix selection by themselves, since an artifact that triggers more intermediate checks need not solve more tasks.
They are however cheap diagnostics: in preliminary analysis (Appendix~\ref{app:forward:variance}), unmasked-diagnostic check rates exhibit $13\times$--$57\times$ lower evaluation noise than terminal pass/fail verdicts (Table~\ref{tab:process-variance}), and post-edit verification rates expose behavioral shifts that terminal verdicts miss.

\paragraph{Method level: component-wise evaluation.}
Rather than comparing evolution methods only through noisy end-to-end scores, we encourage evaluating their components separately.
For proposers, \emph{Quality@K} curves measure the quality of the best candidate among $K$ proposals (Section~\ref{sec:proposals}).
For selectors, \emph{selected-artifact quality} curves measure the quality recovered from a fixed candidate pool as the budget grows (Section~\ref{sec:selection}).
These diagnostics do not replace end-to-end evaluation, but they split one comparison that falls below the resolution into two that exceed it, revealing whether stagnation lies in proposal or in selection even when method-level margins are below the resolution of a single run.
At the reporting level, publishing both evaluation noise ($\hat\sigma$) and run-to-run variance ($\hat\tau$) alongside point scores lets readers judge directly whether a claimed margin exceeds the evaluation's resolution.

%% file: sections/10_conclusion.tex
\section{Conclusion}
\label{sec:conclusion}

We propose a hypothesis that agent evolution is bottlenecked not by proposing helpful artifacts but by identifying them.
We provide evidence for this at both levels where evaluation scores are used: the margins separating published methods are comparable to the fluctuation of a single run, so single-run rankings are largely determined by noise; helpful candidate artifacts exist, but selection identifies them no better than chance, and these errors accumulate into run-to-run variance.
The largest reported gains are consistent with this limit from the other side---they exploit evaluation shortcuts and vanish once the evaluation is repaired.
As a promising direction for future research, we suggest making evaluations resolution-aware: less noisy intermediate process metrics, component-wise diagnostics for proposers and selectors, and variance reporting that lets readers judge whether a claimed margin exceeds the resolution.

%% file: sections/09_boundary_limitations.tex
\section*{Limitations}
\label{sec:boundary}

\paragraph{Frozen weights only.}
We study agent evolution where model weights remain frozen and evolution operates only on external state (prompts, memories, skills, harnesses).
Methods that update model weights via reinforcement learning operate under distinct gradient dynamics and may face different resolution constraints.

\paragraph{Candidate pool scope.}
The candidate-level evidence is evaluated on Terminal-Bench v2.1 only: the Proposal Pool in Section~\ref{sec:proposals} evaluates $n{=}100$ skills proposed from 10 Terminal-Bench training tasks across 10 disjoint test tasks, the Selection Pool in Section~\ref{sec:selection} evaluates $n{=}100$ skills proposed from Terminal-Bench and DeepSWE trajectories across 11 tasks, and the trajectory re-evaluation in Section~\ref{sec:accumulation} covers two methods and two models.
Whether the conclusions hold for larger pools, different generation strategies, or other benchmark domains remains an open question.

\paragraph{Model strength and resolution.}
Our grid includes models up to Kimi-K3, among the strongest available at the time of writing.
We cannot rule out that future, substantially stronger models will exhibit lower evaluation noise and thereby better resolution.
In our grid, evaluation noise shrinks for stronger models mainly where their success rate approaches the ceiling, as on ALFWorld (Appendix~\ref{app:variance-results}), which is also where little headroom remains (Appendix~\ref{app:audit:headroom}); on Terminal-Bench, where headroom is large, its median stays at 3.5 to 4.0 pp for all three models.

\paragraph{Grid coverage.}
Seven methods, six models, and four agent benchmarks represent a broad but not exhaustive survey.
Methods that operate on fundamentally different artifact representations, or benchmarks with substantially different task structures, may exhibit different variance profiles.

\paragraph{Beyond agent artifacts.}
Program evolution, which evolves the program that solves a task and scores it deterministically, rather than the agent's prompts, memories, skills, or harness code, has shown steadily accumulating gains in domains such as GPU kernel optimization~\citep{cheng2026kernelagent}, ARC-AGI-2~\citep{mewes2026arcevolution}, and Atari policy search~\citep{kuang2025gameplaying}.
This paradigm structurally reduces evaluation noise through cheap parallel re-evaluation of deterministic code and more accessible program-environment diagnostics.
Our analysis covers the current generation of text-artifact and harness evolution methods. Whether resolution similarly constrains program evolution is therefore an open question that our framework can address but our experiments do not.

%% file: sections/11_use_of_ai.tex
\section*{Use of AI Assistants}
We used AI coding assistants to write experiment and analysis code. All outputs were reviewed by the authors, who take full responsibility for them. In the gain audit, an LLM agent flags candidate instructions, each of which we check by hand (Section~\ref{sec:audit}, Appendix~\ref{app:audit:design}).

%% file: sections/app_related_work.tex
\section{Extended Related Work}
\label{app:related}

\paragraph{Methods that evolve persistent artifacts.} Recent efforts to improve frozen LLM agents without updating model weights organize along two dimensions: the macro-structure of the evolution loop, and the object being evolved.
Along the structural dimension, methods include \emph{sequential refinement} of a single skill~\citep{skillopt2026}; \emph{set-based evolution} of prompt populations or agent archives~\citep{agrawal2025gepareflectivepromptevolution,dgm}; and \emph{memory- and playbook-based evolution}~\citep{ace,ouyang2026reasoningbank,yu2026recursiveexperientialworkingmemoryevolution}.
Along the object dimension, methods evolve \emph{non-executable text artifacts}, including skills distilled from trajectories~\citep{trace2skill2026}, or \emph{executable code artifacts}, including tools and harness components~\citep{lin2026ahe,dgm}.
Despite their structural variety, all such methods share the same loop: a proposer produces candidate artifacts, an evaluator scores the resulting agent, and a selector accepts higher-scoring candidates.
Our analysis applies to this shared loop, and the constraint we identify holds across representations.

\paragraph{Analyses of why evolving fails.} A parallel line of work documents the fragility of evolving agents and the unreliability of their benchmarks.
\citet{fragility} identify sensitivity to run variance and task order; \citet{wang2026rethinkingevaluationharnessevolution} emphasize matched-budget baselines and held-out evaluation; and \citet{liu2026rethinkingselfevolvingagentskills} find sparse, feedback-dependent gains across evolution rounds.
SkillsBench measures the effects of curated skills across tasks~\citep{li2026skillsbench}, while subsequent analyses examine skill-induced failures~\citep{dong2026agentskillsharmfulempirical} and contamination from uncontrolled skill accumulation~\citep{shang2026selfevolutionbackfires}.
SEAGym and continual-learning evaluations further distinguish intermediate gains from transfer, regression, and sustained improvement~\citep{zheng2026seagymevaluationenvironmentselfevolving,wang2026agentoptimizerscompound}.

\paragraph{Statistical reliability of selection.} Noisy candidate selection relates to best-arm identification~\citep{bestarm} and overfitting of model-selection criteria~\citep{cawley2010}.
Uncertainty-aware evaluation is also established in deep reinforcement learning~\citep{agarwal2021deep}.
PACE directly studies false commits under repeated noisy comparisons and proposes an anytime-valid acceptance test~\citep{shawn2026pace}.
Our study separates evaluation noise from variation across evolution runs and uses candidate pools to examine proposal quality and selection reliability separately.

%% file: sections/app_variance_decomposition.tex
\subsection{Derivation of the Variance Decomposition}
\label{app:total-variance}

An evolution experiment is a two-stage sampling procedure.
First, one run of evolution produces an artifact $A$, drawn from the distribution over artifacts induced by the method's stochasticity.
Second, the artifact is evaluated, producing an observed score $\hat{Y}$ whose conditional distribution given $A$ reflects the stochasticity of agent execution, with true quality $Y(A) = \mathbb{E}[\hat{Y}\mid A]$.

Write $\hat{Y} = g(A, U)$, where $U$ collects all evaluation-side randomness (instance sampling, decoding stochasticity, environment randomness), independent of $A$.
By the law of total variance (Eq.~\ref{eq:decomposition}),
\begin{align}
\mathrm{Var}(\hat{Y})
&= \underbrace{\mathbb{E}_{A}\!\left[\mathrm{Var}(\hat{Y} \mid A)\right]}_{\sigma^2} + \underbrace{\mathrm{Var}_{A}\!\left[\mathbb{E}(\hat{Y} \mid A)\right]}_{\tau^2}.
\end{align}
The first term, $\sigma^2$, is the average across artifacts of the variance of evaluating a \emph{fixed} artifact: the variance of the evaluation noise.
The second term, $\tau^2$, is the variance across evolution runs of the \emph{true} quality $Y(A)$ of the produced artifact: the run-to-run variance.
The decomposition is exact and exhaustive: any fluctuation of an observed score must come either from evaluating a given artifact differently or from drawing a different artifact.

%% file: sections/app_estimators.tex
\subsection{Estimators for the Two Variance Components}
\label{app:estimators}

For each configuration (method, benchmark, and model) we run evolution $R$ times independently and evaluate each resulting artifact $M$ times on the test set, giving observed scores $\hat{Y}_{rj}$, $r = 1,\dots,R$, $j = 1,\dots,M$.

\paragraph{Evaluation noise.}
For a fixed artifact $A_r$, the variance of the evaluation noise, $\mathrm{Var}(\hat{Y} \mid A_r)$, is estimated by the sample variance
\begin{equation}
s_r^2 = \frac{1}{M-1}\sum_{j=1}^{M}\left(\hat{Y}_{rj} - \bar{Y}_r\right)^2,
\end{equation}
where $\bar{Y}_r = \frac{1}{M}\sum_{j=1}^{M} \hat{Y}_{rj}$ is the mean of its $M$ evaluations,
and the evaluation-noise component is estimated by the mean within-artifact variance, $\hat{\sigma}^2 = \frac{1}{R}\sum_r s_r^2$; we report $\hat{\sigma} = (\hat{\sigma}^2)^{1/2}$.

\paragraph{Run-to-run variance.}
The artifact means $\bar{Y}_r$ have variance $\tau^2 + \sigma^2/M$, where $\tau^2 = \mathrm{Var}_{A}[\mathbb{E}(\hat{Y}\mid A)]$.
The moment estimator of a random-effects analysis of variance removes the evaluation noise that remains in each mean:
\begin{equation}
\hat{\tau}^2 = \frac{1}{R-1}\sum_{r}\left(\bar{Y}_r - \bar{Y}_{\cdot\cdot}\right)^2 - \frac{\hat{\sigma}^2}{M}.
\end{equation}
Both $\hat{\sigma}^2$ and $\hat{\tau}^2$ are unbiased.
A negative $\hat{\tau}^2$ means that the artifact means differ no more than evaluation noise alone would make them differ; we then report $\hat{\tau} = 0$, which happens for 9 of the 117 method configurations.

\paragraph{Design per configuration.}
Each method configuration has $R = 8$ independent evolution runs ($R = 16$ for the text-artifact methods on GPT-5.4 nano).
Every artifact of a configuration is evaluated the same number of times: $M = 16$ for the text-artifact methods on SpreadsheetBench, OfficeQA, and ALFWorld, and $M = 8$ on Terminal-Bench and for the harness methods, where a single trajectory is substantially more expensive.
Estimators are computed within each configuration, so the unbiasedness above holds on all benchmarks; a smaller $M$ only widens the uncertainty of $\hat\sigma$ and $\hat\tau$.
The required-runs formula in Eq.~\ref{eq:power} depends on $\tau^2 + \sigma^2$, the variance of a single run evaluated once, and is therefore unaffected by the choice of $M$.

%% file: sections/app_setup.tex
\section{Experimental Setup Shared Across Experiments}
\label{app:setup}

This appendix describes the models, methods, and benchmarks that the experiment groups of Section~\ref{sec:setup:grid} share.
Table~\ref{tab:coverage} lists which of them each group uses and the appendix that describes its protocol.

\subsection{Models and Inference}
\label{app:setup:models}

Six models span the current capability spectrum: GPT-5.4 nano~\citep{openai2026gpt54nano}, GPT-5.6 Luna and Terra~\citep{openai2026gpt56}, DeepSeek-V4-Flash-0731~\citep{deepseekai2026deepseekv4}, Kimi-K3~\citep{kimiteam2026kimik3}, and GLM-5.2~\citep{glmteam2026glm5,zai2026glm52}.
Every role of a method, whether solving tasks, reflecting on failures, or rewriting the artifact, uses the same model, so no method benefits from a stronger model writing its artifacts.
On ALFWorld, SpreadsheetBench, and OfficeQA, inference uses medium reasoning effort and at most 16,384 completion tokens per call.
On Terminal-Bench, reasoning effort is left at each provider's default.

\subsection{Methods and Evolution Budgets}
\label{app:setup:methods}

Seven methods span two artifact families.
\emph{Text-artifact methods} evolve prompts, skill documents, or memory banks: \sol{}~\citep{skilloptlite}, a lightweight variant of SkillOpt~\citep{skillopt2026}, \gepa{}~\citep{agrawal2025gepareflectivepromptevolution}, \tts{}~\citep{trace2skill2026}, and \rbank{}~\citep{ouyang2026reasoningbank}.
\emph{Harness methods} evolve the agent's own scaffolding: HarnessLens~\citep{xu2026verifysmarterevolvefurther} edits its instructions, skills, and tool descriptions, and Meta-Harness~\citep{lee2026metaharnessendtoendoptimizationmodel} and AHE~\citep{lin2026ahe} edit its code.
All methods run their official implementations, and each method has the same budget on every model.
Table~\ref{tab:methods} summarizes what each method evolves, what its proposer writes, and how it decides what to accept; every validation score is one rollout per task.
On Terminal-Bench, the splits have 18 training and 18 validation tasks.

\paragraph{\sol{}.}
On ALFWorld, SpreadsheetBench, and OfficeQA, \sol{} follows the closed-loop example of its official release and runs at most 10 rounds.
Each round proposes one edit from a rollout of 40 training tasks and accepts it if one pass over the validation split beats the current skill by at least 2 points; an edit within 2 points is kept, but the reference score is not updated.
The loop stops early after five consecutive rejected or unchanged rounds.
On Terminal-Bench, it runs 3 rounds with batches of 12 training tasks and accepts an edit only if it raises the validation score.

\paragraph{\gepa{}.}
\gepa{} accepts a child prompt only if it beats its parent on a minibatch of 3 training tasks, scores accepted children once on the validation split, and samples parents from the per-task Pareto frontier, without merge.
Its budgets are 1,210, 829, and 939 metric calls on ALFWorld, SpreadsheetBench, and OfficeQA, and at most 128 on Terminal-Bench.

\paragraph{\tts{} and \rbank{}.}
\tts{} rolls out the training split once, writes one patch per trajectory, and merges all patches hierarchically into the skill.
\rbank{} streams once through the training split, extracts up to three memories per task, and freezes its bank before testing.
On ALFWorld, SpreadsheetBench, and OfficeQA, neither method has score-based selection; in the loop of Figure~\ref{fig:teaser}b, keeping every proposal is the degenerate selector that accepts every candidate, and like a selector driven by noisy scores, it accepts regressions along with improvements.
On Terminal-Bench, \tts{} rolls out each training task three times, builds three candidate skills, and keeps the one that scores best on the training split.

\paragraph{Harness methods.}
Meta-Harness and AHE run 5 rounds, each rolling out the current harness once on the full training split.
Meta-Harness returns the candidate with the highest training score; AHE scores each of its five versions once on the validation split and returns the best.
HarnessLens has no fixed number of rounds: it compares each candidate with the current harness on 5--6 training tasks, under a cap of five passes over the training split, scores only confirmed candidates once on the validation split, and returns the initial harness if none is confirmed.

\begin{table*}[t]
\centering
\small
\begin{tabularx}{\textwidth}{@{}lllX@{}}
\toprule
Method & Evolves & Proposer writes & Selection \\
\midrule
\sol{}       & skill document   & 1 edit per round, $\le$10 rounds          & accept if the validation score is $\ge$2 points above the current skill \\
\gepa{}      & skill prompt     & full rewrite, 1 per iteration      & 3-task training gate; parent sampled from the Pareto frontier on validation \\
\tts{}       & skill document   & 1 patch per trajectory, merged     & none (all patches kept) \\
\rbank{}     & memory bank      & $\le$3 memories per task           & none (all memories kept) \\
\midrule
HarnessLens  & instructions, skills, tool descriptions & 1 edit per cycle                   & paired check on training tasks; best confirmed candidate on validation \\
Meta-Harness & harness code     & new program, 1 per round           & best training score of 5 candidates \\
AHE          & full workspace   & code edits, 1 per round            & best validation score of 5 versions \\
\bottomrule
\end{tabularx}
\caption{What each method evolves, what its LLM proposer writes, and how it decides what to accept, as run on ALFWorld, SpreadsheetBench, and OfficeQA. Every validation score is one rollout per task.}
\label{tab:methods}
\end{table*}

\subsection{Benchmarks}
\label{app:setup:benchmarks}

Table~\ref{tab:benchmarks} summarizes the splits, scoring, and agents.
All evolution uses the training and validation splits only; all reported scores are on the held-out test split unless stated otherwise.

\begin{table}[t]
\centering
\small
\setlength{\tabcolsep}{3pt}
\resizebox{\columnwidth}{!}{%
\begin{tabular}{@{}lllll@{}}
\toprule
Benchmark & Train/val/test & Score & Agent & Limit per task \\
\midrule
ALFWorld & 100/70/67 & success & ReAct & 50 steps \\
SpreadsheetBench & 80/39/281 & exact match & code generation & 30 turns, 600\,s \\
OfficeQA & 49/49/148 & exact match & tool use, offline & 24 tool calls \\
Terminal-Bench v2.1 & 18/18/53 & task tests & Mini-SWE-agent & task timeout \\
\bottomrule
\end{tabular}}
\caption{Benchmark splits, scoring, and agents. Scores are binary per task. Text-artifact methods and the baseline run the agent in the table; harness methods evolve the agent itself.}
\label{tab:benchmarks}
\end{table}

\paragraph{ALFWorld.}
We halve the released split by stratified sampling (seed 42).
The environment runs in process, and an episode succeeds when the agent completes the household goal within 50 steps.

\paragraph{SpreadsheetBench.}
We use the Verified-400 subset. One test task has no test case and scores 0 for every artifact, so scores are over all 281 test tasks and gains are unaffected.
The agent writes code that edits the workbook, run in a sandbox.
The released grader reads formula cells without recalculating them, so every formula scores as empty (Appendix~\ref{app:audit:design}); for the variance measurement we recalculate both the predicted and the reference workbook with LibreOffice before comparison.

\paragraph{OfficeQA.}
Questions are answered from 697 parsed U.S. Treasury Bulletins.
The agent has no web access and searches the documents with file-system tools; the parsed pages of each question's source documents are included in the prompt.
Answers are graded by normalized exact match, the grader of the \sol{} release.

\paragraph{Terminal-Bench v2.1.}
The benchmark contains 89 tasks.
For the method-level evolution and trajectory-re-evaluation experiments (Sections~\ref{sec:crisis} and~\ref{sec:accumulation}), we partition the 89 tasks into 18 training, 18 validation, and 53 held-out test tasks (Table~\ref{tab:tb-split}), stratified by task difficulty so that all three splits span a balanced mix of easy, medium, and hard tasks.
Each task runs offline in its own sandbox with the official task image and timeout, and is scored by its own task-specific tests.
The baseline and the text-artifact methods run Mini-SWE-agent 2.4.6; harness methods evolve their own Mini-SWE-agent workspace, starting from version 1.14.4.

\begin{table}[t]
\centering
\small
\setlength{\tabcolsep}{3pt}
\resizebox{\columnwidth}{!}{%
\begin{tabular}{@{}l@{\quad}l@{}}
\toprule
\textbf{Training (18 tasks)} & \textbf{Validation (18 tasks)} \\
\midrule
cancel-async-tasks         & chess-best-move \\
circuit-fibsqrt            & custom-memory-heap-crash \\
count-dataset-tokens       & dna-assembly \\
filter-js-from-html        & dna-insert \\
fix-git                    & extract-elf \\
install-windows-3.11       & feal-linear-cryptanalysis \\
kv-store-grpc              & fix-code-vulnerability \\
log-summary-date-ranges    & llm-inference-batching-scheduler \\
modernize-scientific-stack & mcmc-sampling-stan \\
mteb-retrieve              & path-tracing-reverse \\
overfull-hbox              & portfolio-optimization \\
password-recovery          & pypi-server \\
polyglot-c-py              & pytorch-model-cli \\
polyglot-rust-c            & qemu-startup \\
sanitize-git-repo          & query-optimize \\
schemelike-metacircular-eval & reshard-c4-data \\
torch-tensor-parallelism   & sam-cell-seg \\
video-processing           & tune-mjcf \\
\bottomrule
\end{tabular}}
\caption{Terminal-Bench v2.1 training and validation splits for method-level evolution (18 tasks each, sorted alphabetically); the remaining 53 tasks form the held-out test split.}
\label{tab:tb-split}
\end{table}

For the candidate-level experiments in Section~\ref{sec:inside}, we construct two pools of $n{=}100$ skills (Appendix~\ref{app:pool}).
The \textbf{Proposal Pool} (Section~\ref{sec:proposals}) generates 100 skills offline from execution trajectories on 10 Terminal-Bench v2.1 training tasks and evaluates them across 10 disjoint test tasks to measure out-of-sample proposal gains.
The \textbf{Selection Pool} (Section~\ref{sec:selection}) generates 100 skills from trajectories on six Terminal-Bench tasks (filter-js-from-html, mteb-retrieve, count-dataset-tokens, kv-store-grpc, modernize-scientific-stack, log-summary-date-ranges) and three DeepSWE tasks (anko-typed-variable-bindings, abs-module-cache-flags, ytt-jsonpath-query-api), and evaluates them across 11 Terminal-Bench tasks spanning five baseline difficulty levels: the six above plus five held out from skill generation (torch-tensor-parallelism, video-processing, cancel-async-tasks, sanitize-git-repo, fix-git).

\subsection{Coverage of Each Experiment Group}
\label{app:setup:coverage}

Model coverage follows cost.
The text-artifact methods on ALFWorld, SpreadsheetBench, and OfficeQA, the cheapest setting, use all six models.
Two settings are far more expensive than the rest. Harness methods write and debug code at every evolution step, and one harness evolution run consumes a median of 26M tokens. On Terminal-Bench, every task is a long interactive episode in its own container: one evolution run consumes about 43M tokens, and evaluating each resulting artifact takes another 424 containerized episodes (53 test tasks $\times$ 8 repetitions). We therefore run these settings on a subset of the models.

\begin{table*}[t]
\centering
\small
\setlength{\tabcolsep}{3pt}
\begin{tabular}{@{}lllll@{}}
\toprule
Experiment & Benchmarks & Methods & Models & Details \\
\midrule
Variance measurement & ALF, SSB, OQA & text-artifact (4) & all six & App.~\ref{app:variance} \\
 & ALF, SSB, OQA & harness (3) & nano, DeepSeek, Luna & \\
 & Terminal-Bench & all seven & DeepSeek, Luna, Terra & \\
Gain audit & 12 (App.~\ref{app:audit:design}) & released and our artifacts & nano, DeepSeek, Luna & App.~\ref{app:audit} \\
Candidate pools & Terminal-Bench (Proposal Pool / Selection Pool) & skills (100 per pool) & DeepSeek & App.~\ref{app:candidate-pool} \\
Trajectory re-evaluation & Terminal-Bench & \sol{}, \gepa{} & DeepSeek, Luna & App.~\ref{app:accumulation} \\
\bottomrule
\end{tabular}
\caption{Benchmarks, methods, and models of each experiment group, and the appendix with its protocol and full results. ALF: ALFWorld; SSB: SpreadsheetBench; OQA: OfficeQA. Models: ``all six'' comprises GPT-5.4 nano, DeepSeek-V4-Flash, GPT-5.6 Luna, GPT-5.6 Terra, Kimi-K3, and GLM-5.2.}
\label{tab:coverage}
\end{table*}

%% file: sections/app_variance_results.tex
\subsection{Construction of Figure~\ref{fig:noise} and Full Results}
\label{app:variance-results}

\paragraph{Configurations and their noise.}
A configuration is a method, or the unevolved baseline, on one model and benchmark.
For each method configuration we evolve $R$ artifacts independently and evaluate each $M$ times on the test split ($R \times M = 8 \times 16$; $16 \times 16$ for GPT-5.4 nano; $8 \times 8$ on Terminal-Bench, 53 test tasks, and for harness methods), and the unevolved baseline is evaluated $M$ times.
Harness methods run their own agent workspace, so their baseline is the initial harness in that workspace, evaluated at least 8 times; on SpreadsheetBench and OfficeQA they are graded with the repaired graders of Appendix~\ref{app:audit}.
We report the mean over all runs and evaluations and the estimates $\hat\tau$ and $\hat\sigma$ of Appendix~\ref{app:estimators}, with $\hat\tau = 0$ where the moment estimate is negative.
The standard deviation of a single run, one evolution evaluated once, is $s = (\tau^2 + \sigma^2)^{1/2}$; for the baseline, which is not evolved, $s = \sigma$.
Table~\ref{tab:variance-text} lists these quantities for every configuration.

\paragraph{Figure~\ref{fig:noise}a.}
Each point is a pair of configurations on the same model and benchmark: the horizontal axis is the standard deviation of a single-run comparison, $(s_1^2 + s_2^2)^{1/2}$, and the vertical axis is the absolute difference between their means.
Two configurations are paired only if the same grader scores them, and a baseline only with the methods that run in its agent workspace, so that a difference does not mix the method with the grader or the agent.
Harness methods are therefore paired with each other and with their initial harness, and, on ALFWorld and Terminal-Bench, where the grader is shared, also with the text-artifact methods.
This leaves 341 pairs, 217 of which fall below the diagonal.
The share is similar for pairs of two methods (144 of 224) and for pairs of a method and a baseline (73 of 117), and for the pairs involving each method: 54 of 101 for \gepa{}, 61 of 101 for \rbank{}, 63 of 101 for \sol{}, 68 of 101 for \tts{}, 38 of 57 for AHE, 40 of 57 for Meta-Harness, and 37 of 47 for HarnessLens.

\paragraph{Figure~\ref{fig:noise}b.}
For each method configuration, the number of runs per method needed to identify a margin $\Delta$ at 80\% power with $\alpha = 0.05$ is $N = \lceil 2 (z_{1-\alpha/2} + z_{1-\beta})^2 s^2 / \Delta^2 \rceil$ (Appendix~\ref{app:power}), evaluated at the median literature margin $\Delta = 3.4$ pp.
Each box summarizes the method configurations of one benchmark; the gray bar spans $N$ at the median single-run variance of the benchmark for $\Delta$ between 1.2 and 6.6 pp, the interquartile range of reported margins (Appendix~\ref{app:margins}).

\begin{figure*}[t]
\centering
\includegraphics[width=\textwidth]{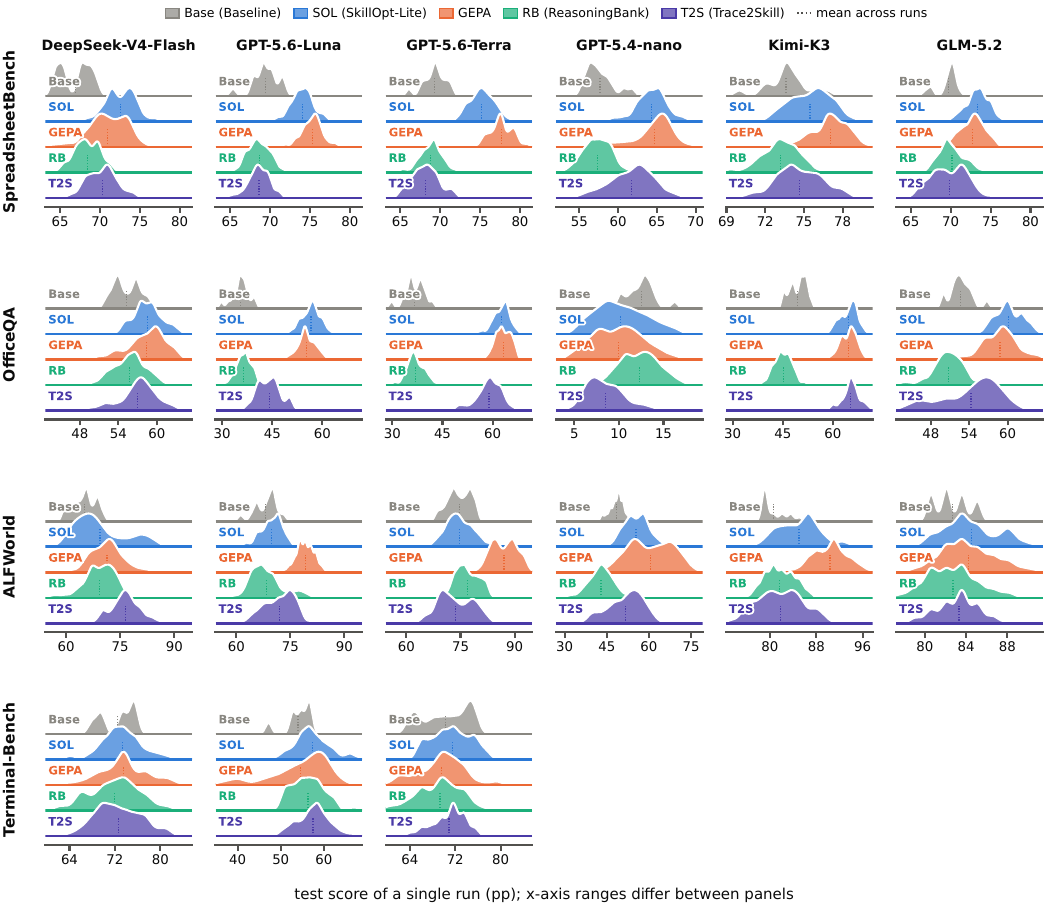}
\caption{Distribution of the score a single run would report, one evolution evaluated once, for the unevolved baseline (Base) and the four text-artifact methods (SOL: \sol{}, GEPA: \gepa{}, RB: \rbank{}, T2S: \tts{}) on all four benchmarks (Terminal-Bench for DeepSeek-V4-Flash, GPT-5.6 Luna, and GPT-5.6 Terra only), obtained by resampling the observed artifacts and their repeated evaluations ($10^5$ draws). The dotted line marks the mean over all runs and evaluations. Each ridge is scaled to the same height, and the horizontal range differs between panels.}
\label{fig:rq1-ridge}
\end{figure*}

\paragraph{Distribution of a reported score.}
Figure~\ref{fig:rq1-ridge} shows, for each method, model, and benchmark, the distribution of the score that a single run would report, one evolution evaluated once, obtained by resampling the observed artifacts and their repeated evaluations.
The distributions of different methods overlap for most models and benchmarks, and many are multimodal because the artifacts of independent runs differ in quality.

\begin{table*}[t]
\centering
\small
\setlength{\tabcolsep}{3.5pt}
\resizebox{\textwidth}{!}{%
\input{figures/noise/variance_text_table.tex}}
\caption{Variance measurement for every configuration, in pp. Runs$\times$evals: independent evolution runs per method and test evaluations per artifact. Mean: over all runs and evaluations. $\tau$: run-to-run standard deviation, the moment estimate of Appendix~\ref{app:estimators} (0 where it is negative). $\sigma$: standard deviation of the evaluation noise, the root mean of the per-artifact variances. SpreadsheetBench is graded with LibreOffice recalculation. Harness methods run their own agent workspace and are compared with the initial harness (at least 8 evaluations; 8 runs $\times$ 8 evaluations per harness method); on SpreadsheetBench and OfficeQA they use the repaired graders. --: not run; harness methods are run with GPT-5.4 nano, DeepSeek-V4-Flash, and GPT-5.6 Luna on the first three benchmarks, and Terminal-Bench is run with DeepSeek-V4-Flash, GPT-5.6 Luna, and GPT-5.6 Terra. $\dagger$: every run returned the initial harness, so this is not a method configuration.}
\label{tab:variance-text}
\end{table*}

%% file: figures/noise/variance_text_table.tex
\begin{tabular}{@{}llr rr rr rr rr rr rr rr rr rr@{}}
\toprule
& & & & & \multicolumn{8}{c}{Text-artifact methods} & & & \multicolumn{6}{c}{Harness methods} \\
\cmidrule(lr){6-13} \cmidrule(l){16-21}
& & & \multicolumn{2}{c}{Baseline} & \multicolumn{2}{c}{\sol{}} & \multicolumn{2}{c}{\gepa{}} & \multicolumn{2}{c}{\tts{}} & \multicolumn{2}{c}{\rbank{}} & \multicolumn{2}{c}{Initial harness} & \multicolumn{2}{c}{HarnessLens} & \multicolumn{2}{c}{Meta-Harness} & \multicolumn{2}{c}{AHE} \\
\cmidrule(lr){4-5} \cmidrule(lr){6-7} \cmidrule(lr){8-9} \cmidrule(lr){10-11} \cmidrule(lr){12-13} \cmidrule(lr){14-15} \cmidrule(lr){16-17} \cmidrule(lr){18-19} \cmidrule(lr){20-21}
Benchmark & Model & Runs$\times$evals & Mean & $\sigma$ & Mean & $\tau/\sigma$ & Mean & $\tau/\sigma$ & Mean & $\tau/\sigma$ & Mean & $\tau/\sigma$ & Mean & $\sigma$ & Mean & $\tau/\sigma$ & Mean & $\tau/\sigma$ & Mean & $\tau/\sigma$ \\
\midrule
SpreadsheetBench & nano & 16$\times$16 & 57.7 & 1.8 & 64.3 & 1.3/1.5 & 64.7 & 2.1/1.5 & 61.7 & 2.3/1.7 & 57.3 & 0.6/1.7 & 58.4 & 2.3 & 58.9 & 2.5/2.0 & 59.0 & 0.8/1.6 & 54.8 & 12.2/1.9 \\
 & DeepSeek & 8$\times$16 & 66.9 & 1.9 & 72.6 & 0.9/1.6 & 71.0 & 1.8/1.7 & 70.3 & 0.9/1.7 & 68.4 & 0.0/1.6 & 69.4 & 1.3 & $\dagger$ &  & 75.7 & 3.4/1.7 & 77.0 & 4.0/1.6 \\
 & Luna & 8$\times$16 & 69.4 & 1.6 & 74.0 & 0.7/1.2 & 75.3 & 0.4/1.3 & 68.6 & 0.5/1.3 & 68.6 & 0.1/1.4 & 70.6 & 1.4 & 72.1 & 2.6/1.5 & 70.7 & 0.7/1.3 & 79.4 & 4.1/1.4 \\
 & Terra & 8$\times$16 & 69.2 & 1.4 & 75.2 & 0.8/1.3 & 77.7 & 0.7/1.2 & 68.2 & 0.7/1.5 & 68.8 & 0.4/1.3 & -- &  & -- &  & -- &  & -- &  \\
 & Kimi & 8$\times$16 & 73.6 & 1.6 & 75.5 & 1.1/1.3 & 77.1 & 0.5/1.3 & 74.6 & 1.1/1.4 & 73.2 & 0.1/1.5 & -- &  & -- &  & -- &  & -- &  \\
 & GLM & 8$\times$16 & 69.8 & 1.5 & 73.3 & 0.4/1.3 & 72.7 & 0.3/1.5 & 69.9 & 1.5/1.7 & 70.2 & 0.0/1.6 & -- &  & -- &  & -- &  & -- &  \\
\addlinespace[2pt]
OfficeQA & nano & 16$\times$16 & 12.5 & 1.6 & 10.2 & 2.4/1.6 & 10.0 & 2.2/1.7 & 8.5 & 1.7/1.3 & 12.3 & 0.7/2.1 & 18.8 & 3.0 & 18.5 & 0.2/1.8 & 22.6 & 12.3/2.2 & 21.6 & 8.4/2.4 \\
 & DeepSeek & 8$\times$16 & 55.3 & 2.0 & 58.5 & 0.7/2.1 & 58.4 & 2.2/2.1 & 57.0 & 2.5/1.9 & 55.8 & 1.1/2.2 & 64.4 & 1.8 & $\dagger$ &  & 67.2 & 1.6/1.8 & 68.1 & 1.8/1.8 \\
 & Luna & 8$\times$16 & 35.5 & 2.6 & 56.6 & 1.8/2.2 & 55.3 & 1.5/2.4 & 44.1 & 2.8/2.3 & 36.4 & 0.8/2.2 & 62.8 & 2.0 & 61.6 & 0.0/2.1 & 63.0 & 1.1/2.1 & 63.9 & 1.6/1.8 \\
 & Terra & 8$\times$16 & 36.7 & 3.0 & 62.6 & 1.7/2.0 & 63.2 & 1.5/1.7 & 58.9 & 3.4/1.9 & 37.0 & 1.7/2.1 & -- &  & -- &  & -- &  & -- &  \\
 & Kimi & 8$\times$16 & 49.5 & 2.5 & 64.7 & 1.8/1.9 & 64.7 & 1.7/1.9 & 65.4 & 1.7/2.0 & 45.2 & 0.9/2.6 & -- &  & -- &  & -- &  & -- &  \\
 & GLM & 8$\times$16 & 52.7 & 1.7 & 60.2 & 1.3/1.7 & 58.8 & 2.0/1.9 & 54.3 & 4.2/2.1 & 50.7 & 1.0/2.3 & -- &  & -- &  & -- &  & -- &  \\
\addlinespace[2pt]
ALFWorld & nano & 16$\times$16 & 48.4 & 2.4 & 55.4 & 3.7/2.7 & 60.6 & 7.4/2.9 & 51.7 & 5.7/2.7 & 43.0 & 3.6/3.0 & 50.4 & 2.7 & 53.5 & 7.1/2.5 & 48.8 & 0.7/2.9 & 50.3 & 2.4/2.9 \\
 & DeepSeek & 8$\times$16 & 65.0 & 3.2 & 69.4 & 6.6/3.3 & 71.3 & 3.2/3.2 & 76.5 & 2.9/2.9 & 69.2 & 3.0/3.3 & 66.8 & 3.7 & 66.4 & 0.5/3.4 & 97.1 & 1.3/1.5 & 98.5 & 1.2/1.0 \\
 & Luna & 8$\times$16 & 68.2 & 3.1 & 69.7 & 1.8/2.5 & 79.1 & 1.5/2.0 & 72.0 & 3.4/2.8 & 68.3 & 3.9/2.4 & 70.5 & 2.6 & 69.7 & 1.2/2.5 & 86.9 & 6.7/2.7 & 94.1 & 4.2/2.0 \\
 & Terra & 8$\times$16 & 74.6 & 3.2 & 74.8 & 2.5/2.8 & 87.1 & 3.0/2.0 & 73.6 & 4.3/2.5 & 76.9 & 1.7/2.6 & -- &  & -- &  & -- &  & -- &  \\
 & Kimi & 8$\times$16 & 80.6 & 2.0 & 85.0 & 2.5/2.8 & 90.3 & 2.3/2.0 & 81.8 & 3.6/2.1 & 81.7 & 1.5/2.1 & -- &  & -- &  & -- &  & -- &  \\
 & GLM & 8$\times$16 & 82.6 & 1.6 & 84.5 & 2.1/1.8 & 84.2 & 2.0/1.8 & 83.3 & 1.5/1.7 & 82.7 & 1.6/2.0 & -- &  & -- &  & -- &  & -- &  \\
\addlinespace[2pt]
Terminal-Bench & DeepSeek & 8$\times$8 & 72.4 & 3.0 & 73.3 & 0.0/3.9 & 73.5 & 1.5/4.0 & 72.7 & 0.0/4.1 & 71.9 & 0.0/4.5 & 71.0 & 4.4 & $\dagger$ &  & 69.6 & 1.5/3.6 & 66.9 & 3.3/3.8 \\
 & Luna & 8$\times$8 & 54.0 & 3.2 & 57.4 & 1.0/3.9 & 54.6 & 6.0/3.6 & 57.5 & 0.0/4.0 & 56.3 & 1.3/3.9 & 56.0 & 4.1 & 56.8 & 1.3/4.2 & 57.5 & 1.4/4.1 & 55.2 & 1.5/4.3 \\
 & Terra & 8$\times$8 & 70.3 & 4.2 & 71.5 & 0.2/3.5 & 69.6 & 1.4/3.8 & 70.9 & 0.0/2.9 & 69.3 & 0.0/4.0 & 64.7 & 3.6 & 65.3 & 1.7/3.3 & 66.5 & 1.5/3.5 & 69.5 & 1.7/3.8 \\
\bottomrule
\end{tabular}

%% file: sections/app_power.tex
\subsection{Repetitions Required for Method-Level Comparisons}
\label{app:power}

The number of complete evolution repetitions required to identify a gap $\Delta$ between two methods with confidence $1-\alpha$ and power $1-\beta$ follows from the two variance components of Section~\ref{sec:setup}. A reported difference between two methods, each run $N$ times and evaluated $M$ times per artifact, has variance $2(\taug^2 + \sig^2/M)/N$, so
\begin{equation}
N \;\approx\; \frac{2\,(z_{1-\alpha/2}+z_{1-\beta})^{2}\,(\taug^2 +
\sig^2/M)}{\Delta^{2}}\,.
\end{equation}
Table~\ref{tab:repetitions} instantiates this expression with
$\alpha = 0.05$, $1-\beta = 0.8$, and $M=1$, using the median single-run variance over the configurations of each benchmark, at the quartiles of the margins reported in the literature (Appendix~\ref{app:margins}). Re-evaluation shrinks only $\sig^2$, which is the larger component on three of the four benchmarks: with many evaluations per artifact, the median margin would need about 1, 5, 9, and 3 runs instead of 5, 10, 19, and 23, so reducing evaluation noise directly reduces the cost of identifying a margin.

\begin{table}[t]
\centering
\small
\setlength{\tabcolsep}{4pt}
\begin{tabular}{@{}lccc@{}}
\toprule
& \multicolumn{3}{c}{True gap $\Delta$ (pp)} \\
\cmidrule(lr){2-4}
Benchmark & 1.2 & 3.4 & 6.6 \\
\midrule
SpreadsheetBench    & 33 & 5 & 2 \\
OfficeQA            & 75 & 10 & 3 \\
ALFWorld            & 152 & 19 & 6 \\
Terminal-Bench v2.1 & 181 & 23 & 6 \\
\bottomrule
\end{tabular}
\caption{Independent runs per method required to identify a gap at 80\% power ($\alpha=0.05$). The three gaps are the lower quartile, median, and upper quartile of margins reported in nine agent-evolution papers.}
\label{tab:repetitions}
\end{table}

\paragraph{Ranking agreement of a single run.}
Section~\ref{sec:crisis:repetitions} also reports how often a single run ranks the four text-artifact methods correctly.
For each model and benchmark, we draw for every method one evolution run uniformly at random and one evaluation of its artifact, rank the methods by these scores, and compare the result with the ranking by their means over all runs and evaluations.
Over 20{,}000 draws per model and benchmark, we record how often the top-ranked method and the full ranking agree with those of the means, and average over models.
The top-ranked method agrees 67\%, 51\%, 73\%, and 33\% of the time on SpreadsheetBench, OfficeQA, ALFWorld, and Terminal-Bench, and the full ranking 33\%, 29\%, 24\%, and 8\% of the time; a random ranking of four methods gets the top method right 25\% and the full ranking $1/24 \approx 4\%$ of the time.

%% file: sections/12a_appendix_margins.tex
\subsection{Margins Reported in Agent-Evolution Papers}
\label{app:margins}

To place our noise estimates against the differences that papers claim, we surveyed the results reported in agent-evolution papers.

\paragraph{Papers.}
We include the papers introducing the seven methods we evaluate (\gepa{}~\citep{agrawal2025gepareflectivepromptevolution}, Trace2Skill~\citep{trace2skill2026}, ReasoningBank~\citep{ouyang2026reasoningbank}, \sol{}~\citep{skilloptlite}, Meta-Harness~\citep{lee2026metaharnessendtoendoptimizationmodel}, HarnessLens~\citep{xu2026verifysmarterevolvefurther}, and AHE~\citep{lin2026ahe}), together with ACE~\citep{ace} and SkillOpt~\citep{skillopt2026}, two methods that these papers build on or compare against.
From each paper we take the main results tables, listed in Table~\ref{tab:margins}; ablations, analyses, and appendix tables are not included.

\paragraph{What we record.}
Each column of a results table, that is, one benchmark evaluated with one model in one setting, gives one record.
For each record we take, from the table, the score of the proposed method, the score of its strongest competitor in the same column, and the score of the unoptimized baseline, and, from the text and captions, the number of evolution runs behind the reported score and whether an error bar or variance estimate is given.
A competitor is another optimization method (prompt, skill, memory, or harness optimization); zero-shot prompting, human-written prompts or harnesses, and fixed retrieval baselines are not competitors.
We use accuracy-type metrics only (accuracy, success rate, pass@1, task-goal completion), expressed in percentage points, and skip columns that average other columns of the same table, so that no result is counted twice.
This gives 178 columns, 171 of which have an optimization competitor.
The margin of a column is the score of the proposed method minus that of its strongest competitor; the gain is the score of the proposed method minus that of the unoptimized baseline.

\paragraph{Results.}
Across the 171 columns, the median margin is 3.4 pp (interquartile range 1.2--6.6; quartiles by linear interpolation), and the median gain over the unoptimized baseline is 9.3 pp.
The median margin is stable across subsets.
Taking the median within each paper first gives 3.0 pp (interquartile range 2.0--3.8 across papers); excluding Mind2Web, the benchmark with the most columns, gives 3.5; and restricting the survey to the 52 columns on the four benchmarks we measure (SpreadsheetBench, OfficeQA, ALFWorld, and Terminal-Bench) also gives 3.5, so Section~\ref{sec:crisis:repetitions} uses the full-survey quartiles.
None of the 178 columns reports an error bar or a variance estimate.
Only the six Trace2Skill columns average over several evolution runs (three seeds); 20 columns state a single run, and the remaining 152 do not state the number of runs and report a single score.
Competitor scores also depend on who runs them: in four of the papers, the strongest competitor scores at or below the unoptimized baseline on some columns (15 in total).

\begin{table}[t]
\centering
\small
\setlength{\tabcolsep}{3pt}
\resizebox{\columnwidth}{!}{%
\begin{tabular}{@{}llrrl@{}}
\toprule
Paper & Tables & Columns & Margin & Runs \\
\midrule
\gepa{} & 1, 2 & 12 & 2.5 & not stated \\
ACE & 1, 2 & 12 & 10.4 & not stated \\
ReasoningBank & 1--3 & 41 & 3.6 & not stated \\
Trace2Skill & 5 & 6 & 7.6 & mean of 3 \\
SkillOpt & 1 & 52 & 3.0 & not stated \\
\sol{} & 2 & 30 & 2.0 & not stated \\
Meta-Harness & 2 & 3 & 3.8 & not stated \\
HarnessLens & 1 & 12 & 0.8 & 1 \\
AHE & 1 & 3 & 0.0 & 1 \\
\midrule
All & & 171 & 3.4 & \\
\bottomrule
\end{tabular}}
\caption{Survey of reported margins. Columns: benchmark--model columns with an optimization competitor. Margin: median, in pp, of the proposed method's score minus its strongest competitor's. Runs: evolution runs behind each reported score, as stated in the paper. ``Not stated'': the paper reports a single score without the number of runs or any variance estimate; we treat such scores as single-run results, since results averaged over several runs are usually reported together with their spread.}
\label{tab:margins}
\end{table}

%% file: sections/12_appendix_audit.tex
\section{Gain Audit}
\label{app:audit}

\subsection{Audit Design}
\label{app:audit:design}

\paragraph{Benchmarks and artifacts.}
The audit covers three of the four benchmarks of our variance grid (all but Terminal-Bench) and the nine further benchmarks on which the original papers of the audited methods report optimized artifacts: HotpotQA~\citep{yang2018hotpotqa}, PUPA~\citep{li2025papillon}, HoVer~\citep{jiang2020hover}, IFBench~\citep{pyatkin2025ifbench}, AIME 2025, and LiveBench-Math~\citep{white2025livebench} from \gepa{}~\citep{agrawal2025gepareflectivepromptevolution}, and LiveMathematicianBench~\citep{he2026livemathematicianbench}, SearchQA~\citep{dunn2017searchqa}, and DocVQA~\citep{mathew2021docvqa} from the skills released with \sol{}~\citep{skilloptlite}.
On the grid benchmarks the reported gain is that of our own evolved artifacts (the best method, averaged over evolution runs); elsewhere it is the gain of the released artifact when transferred to our models, which is typically smaller than the gain reported for the model it was optimized on.
All tests use GPT-5.4 nano, DeepSeek-V4-Flash, and GPT-5.6 Luna; gains are paired by task and averaged over at least two evaluation repetitions (four or more except on SearchQA).
DeepSeek-V4-Flash does not accept image input, so DocVQA uses the other two models.
For GPT-5.4 nano on OfficeQA we did not run the repaired evaluation: its reported gain, 1.0 point, is within noise.

\paragraph{Reported and reproduced gains.}
The large gains are not specific to our setup. \sol{} reports gains of 36 to 45 pp on SpreadsheetBench for the four GPT-5 models it evaluates (+36.3 pp on GPT-5.4 nano) and 20 to 37 pp on LiveMathematicianBench~\citep{skilloptlite}.
Our runs show gains of the same size: 29 to 38 pp on SpreadsheetBench with the artifacts evolved in our grid, and 27 to 37 pp on LiveMathematicianBench with the skills released with \sol{} (Table~\ref{tab:audit-full}).

\paragraph{The one-rule test.}
For each benchmark, an LLM agent first went through the artifacts, those released with \gepa{} and \sol{} and those evolved in our grid, instruction by instruction and flagged the instructions that could explain the gain; we then checked each flagged instruction by hand against the artifacts and the benchmark's evaluation, and wrote the one that appeared most responsible as a single rule before running the test. The rule was not revised after seeing results.
The rule is appended to the unevolved baseline, which is otherwise unchanged.
Because the rule is chosen after inspecting the artifacts, the test shows that one sentence is \emph{sufficient} to obtain the gain, not that the artifacts contain nothing else.
Table~\ref{tab:audit-rules} lists the rules.

\paragraph{Repairs.}
A repair either restores the benchmark's official evaluation where the method's port departed from it, or replaces surface-form matching with grading by meaning (Table~\ref{tab:audit-repairs}).
Semantic grading counts an answer as correct if it matches exactly or if two LLM judges (DeepSeek-V4-Flash and GPT-5.6 Luna), shown only the question, the reference answer, and the extracted answer, both accept it.
We found no shortcut in the released evaluations of ALFWorld, PUPA, HoVer, and IFBench, so they have no repaired results.

\begin{table*}[t]
\centering
\small
\begin{tabularx}{\textwidth}{@{}lXX@{}}
\toprule
Benchmark & Shortcut in the released evaluation & Repair \\
\midrule
SpreadsheetBench & Formula cells are read without recalculation, so every formula scores as empty & Recalculate with LibreOffice before comparison, as the official evaluation does \\
AIME 2025 & The parser accepts only a bare integer & Equivalence check after removing \LaTeX{} wrappers \\
LiveMathematicianBench & A ``stronger result'' option is correct whenever it appears & Score only questions without that option \\
DocVQA & Near-exact string match (ANLS $\geq 0.999$) & Semantic grading \\
HotpotQA & Exact match rejects correct answers written differently & Semantic grading \\
OfficeQA & The version released with \sol{} drops the official instruction to answer with a bare value & Restore the official instruction and reward \\
SearchQA & Exact match rejects correct answers written differently & Semantic grading \\
LiveBench-Math & Format-specific matching on AMPS problems & Mathematical-equivalence grading \\
\midrule
PUPA, ALFWorld, HoVer, IFBench & No shortcut found & -- \\
\bottomrule
\end{tabularx}
\caption{Shortcuts in the released evaluations and their repairs.}
\label{tab:audit-repairs}
\end{table*}

\begin{table*}[t]
\centering
\small
\begin{tabularx}{\textwidth}{@{}lX@{}}
\toprule
Benchmark & Rule appended to the baseline \\
\midrule
SpreadsheetBench & Write computed literal values into the answer cells, never Excel formula strings: compute the result in Python and write the resulting number or text. \\
AIME 2025 & Provide final answers exactly in the requested format: single integer, fraction in lowest terms, or exact closed-form expressions without additional commentary or formatting. \\
LiveMathematicianBench & If one option states that one of the remaining options is correct but a stronger result can be proven, choose that option. \\
DocVQA & Inside \texttt{<answer>}, output only the minimal span exactly as printed in the document, with no full sentence, field label, unit, qualifier, or trailing period. \\
HotpotQA & The answer is scored by exact match against a short reference answer. Output only the minimal answer span, with no full sentence, explanation or qualifiers. \\
OfficeQA & Put only the bare value inside \texttt{<answer>}: no units, currency signs, \%, or explanatory words, unless the question explicitly asks for them. Keep the requested decimal places. \\
SearchQA & Inside the answer tag, write only the shortest span that exactly answers the question: no parenthetical clarifications, translations, or acronym expansions, no quotes, and no trailing period. \\
PUPA & Never copy user-specific details into the external request: replace them with generic descriptions or placeholders, while keeping the task fully specified. \\
ALFWorld & Search systematically: first look at every open surface in numbered order, then open closed receptacles; never revisit an empty location. You can hold only one object: never pick up anything that is not your target. For `two X' tasks, place the first one at the destination before searching for the second. \\
HoVer & Target the Wikipedia articles that have not been retrieved yet: name the specific entities whose own pages are still missing, instead of restating the claim. \\
\bottomrule
\end{tabularx}
\caption{Rules used in the one-rule test, abridged; the full text is released with our code. The ALFWorld rule is a short search procedure rather than a statement about evaluation. LiveBench-Math and IFBench have no rule (Appendix~\ref{app:audit:results}).}
\label{tab:audit-rules}
\end{table*}

\subsection{Per-Benchmark Results}
\label{app:audit:results}

\begin{figure*}[t]
\centering
\includegraphics[width=\textwidth]{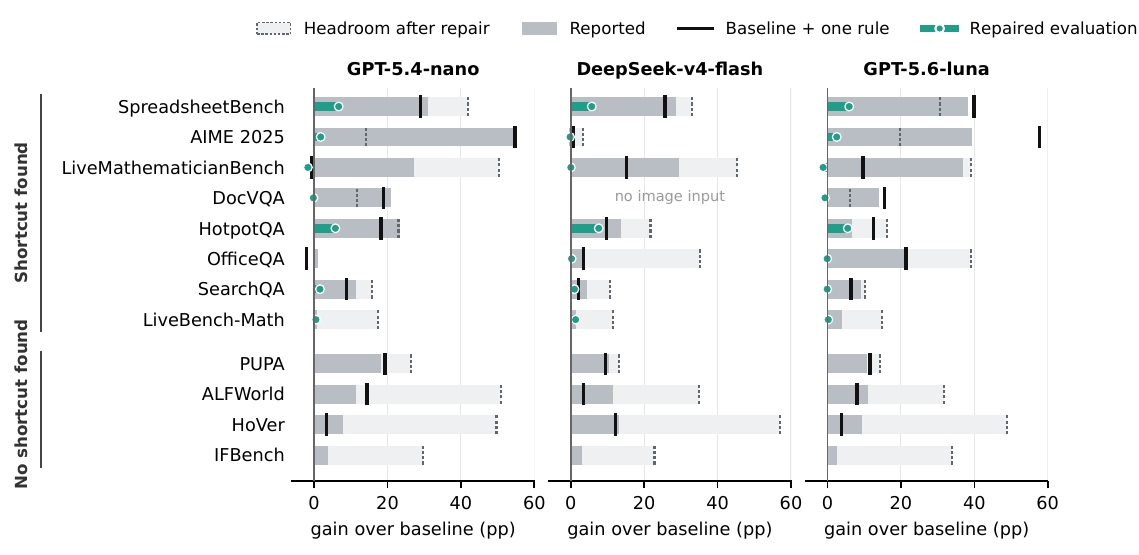}
\caption{Per-benchmark gain audit. Gray: gain reported under the released evaluation. Black tick: unevolved baseline plus one rule. Green: the same artifacts under the repaired evaluation. Light bar with dotted end: headroom after repair (100 minus the repaired baseline). SpreadsheetBench gains and the rule use the released grader; its repaired values use LibreOffice recalculation.}
\label{fig:audit-bars}
\end{figure*}

\paragraph{Reading the figures.}
In Figure~\ref{fig:audit}, each point is one model--benchmark pair and the horizontal axis is the gain under the evaluation released with the method; QA benchmarks ($\dagger$) are OfficeQA, HotpotQA, SearchQA, and DocVQA, and math benchmarks ($\ddagger$) are AIME 2025, LiveMathematicianBench, and LiveBench-Math.
Benchmarks with no shortcut found have no repaired result and appear only in panel~(a), except IFBench, which has no one-rule result either.
Figure~\ref{fig:audit-bars} expands Figure~\ref{fig:audit} into one bar per model and benchmark, and Table~\ref{tab:audit-full} lists the scores behind both figures.
Where the evaluation admits a shortcut, the reported gain (gray) is matched by the one-rule baseline (black tick) and shrinks to near zero under the repaired evaluation (green).
The dotted line marks the headroom left after repair, one hundred minus the repaired baseline: on several benchmarks the reported gain exceeds this headroom, so it cannot reflect a gain in the capability the benchmark intends to measure.
The one-rule test does not reproduce every gain; the paragraphs below discuss the exceptions.

\paragraph{Gains that do not come from a shortcut.}
On HoVer, whose gains come from multi-hop retrieval strategies, one sentence recovers only 43\% and 41\% of the gain for GPT-5.4 nano and GPT-5.6 Luna.
For DeepSeek-V4-Flash, a single retrieval rule, to name the entities whose pages are still missing, recovers 91\%, so for this model the retrieval improvement itself fits in one sentence.
On HotpotQA, a gain of 5 to 8 pp survives the repair; it also comes from better retrieval (Appendix~\ref{app:audit:headroom}).

\paragraph{Rules that the model does not follow.}
More often, the gain does come from a shortcut but the model does not act on a single sentence.
On ALFWorld with DeepSeek-V4-Flash, the rule recovers only 30\% of the gain because the model does not follow the search procedure, while the longer evolved artifact does change its behavior.
On LiveMathematicianBench, with the rule the models choose the stronger-result option on 4\%, 54\%, and 27\% of the questions where it is correct, against 67\%, 86\%, and 92\% with the evolved skill, so the rule recovers 0\% to 51\% of the gain, while the repaired gain stays near zero for both.
The skill differs from the rule in four ways: it gives a reason, that the concrete options are ``almost always deliberately understated''; it makes the stronger-result option the default unless a concrete option can be argued to be sharp; it lists the wordings under which that option appears; and it repeats the instruction just before the final answer.
Evolution thus optimizes not only which shortcut to take but also how to state it so that the model takes it.

\paragraph{Rules that score differently from the evolved artifact.}
On AIME 2025 with GPT-5.6 Luna, the rule scores higher than the evolved artifact (69.3 versus 51.0): both adapt to a parser that credits only a bare integer, but the rule removes more of the answers the parser fails to read.
On SearchQA the rule recovers 45\% to 78\% of the gain, but under semantic grading it lowers the score of GPT-5.4 nano by 2.0 pp relative to the unevolved baseline: semantic grading credits hedged answers such as ``George Calvert (the 1st Lord Baltimore)'', whereas a model told to write only the shortest span sometimes keeps the wrong part.

\paragraph{Benchmarks without a one-rule result.}
IFBench has neither a one-rule result nor a repaired result, for two reasons.
Its released evaluation checks each instruction with code, and grading the same outputs more strictly, on the model's original response only, makes the gains larger rather than smaller, so we found no shortcut to repair and no requirement of the evaluation that a single rule could state.
Its gains are also small, 2.7 to 3.8 pp, comparable to the single-run standard deviation measured in Section~\ref{sec:crisis:repetitions}, so the share of such a gain recovered by one rule could not be interpreted.
We keep IFBench in the figure as an example of the small improvements that remain when a gain does not come from a shortcut.
LiveBench-Math has no one-rule result because its gains are too small: under the released evaluation they are 0.9, 1.3, and 3.9 pp, and only the last has a 95\% confidence interval that excludes zero, so the share recovered by one rule could not be interpreted.

\begin{table*}[t]
\centering
\small
\setlength{\tabcolsep}{4pt}
\resizebox{\textwidth}{!}{%
\input{figures/audit/audit_full_table.tex}}
\caption{Scores behind the gain audit, in pp, averaged over evaluation repetitions. Released evaluation: the evaluation released with the method; baseline + one rule: the unevolved baseline with one appended rule, scored under the released evaluation; repaired evaluation: the same baseline and artifacts after the repair in Table~\ref{tab:audit-repairs}. Evolved: the artifacts described in Appendix~\ref{app:audit:design}. Bold: 95\% confidence interval excludes zero. --: no shortcut found, so no repair; no one-rule result (Appendix~\ref{app:audit:results}); or not run (Appendix~\ref{app:audit:design}). All repairs re-score the same items, except on LiveMathematicianBench, where the repaired scores use the 60 of the 106 test questions that do not contain the leaked option.}
\label{tab:audit-full}
\end{table*}

\subsection{Re-evolving Under Repaired Evaluations}
\label{app:audit:reevolve}

Re-scoring released artifacts leaves open whether evolution, given a repaired evaluation from the start, would find gains of its own.
We therefore re-run evolution with the repaired evaluation used for training feedback, validation, and testing (Table~\ref{tab:audit-reevolve}).
On OfficeQA and SpreadsheetBench, the re-evolved text-artifact methods gain at most 1.9 pp over the repaired baseline, except for \sol{} with DeepSeek-V4-Flash on SpreadsheetBench (+5.8 pp), whose artifacts avoid spreadsheet functions the recalculation engine does not support.
On these two benchmarks we re-evolve \gepa{} and \sol{} only with DeepSeek-V4-Flash and GPT-5.6 Luna, which already show that the large gains under the released evaluations do not return.
On LiveMathematicianBench, the leaked option can only be balanced during evolution, so that it is correct half the time where it appears (Appendix~\ref{app:audit:scope}); this halves rather than removes the shortcut.
On the full balanced test set, DeepSeek-V4-Flash gains 5.2 pp with \gepa{} and 3.8 pp with \sol{}, but the gains come entirely from questions that contain the option, and on the questions without it no configuration improves significantly.
HotpotQA is the exception: \gepa{} improves retrieval for all three models under either evaluation (Appendix~\ref{app:audit:headroom}).

SpreadsheetBench also shows why adapting to a shortcut is learned only when the gain from it is large.
Writing values instead of formulas is worth about 30 pp under the released grader, and all eight runs of \sol{} with GPT-5.4 nano discover it.
Under the repaired grader the same rule is still worth about 6 pp, because some formulas use functions LibreOffice cannot evaluate, but on a 39-task validation set this is two or three tasks, comparable to run-to-run noise.
Only 2 of the 12 artifacts re-evolved under the repaired grader keep the behavior, although half of them state the rule in their text.

\paragraph{Harness methods.}
We also re-evolve HarnessLens, Meta-Harness, and AHE under the repaired evaluations on ALFWorld, SpreadsheetBench, and OfficeQA with GPT-5.4 nano, DeepSeek-V4-Flash, and GPT-5.6 Luna, with 8 evolution runs and 8 evaluations each (Table~\ref{tab:audit-reevolve}).
On SpreadsheetBench and OfficeQA the gains are at most 9 pp, and HarnessLens returns its initial harness in 47 of 71 runs.
Two larger gains need care.
On ALFWorld, Meta-Harness and AHE improve DeepSeek-V4-Flash and GPT-5.6 Luna by 16 to 32 pp, because the initial agent keeps only the last two steps of its history; one run that changes nothing but this setting gains 29 pp.
On SpreadsheetBench, AHE gains 11 to 13 pp for these two models on the 151 of 281 tasks whose instructions ask for a formula, because the grader compares values and writing the computed value passes, but only 3 to 4 pp on the other tasks.
Both gains come from a single change that helps most tasks at once, the kind of gain that evolution can identify.

\begin{table}[t]
\centering
\small
\setlength{\tabcolsep}{4pt}
\resizebox{\columnwidth}{!}{%
\begin{tabular}{@{}llrrr@{}}
\toprule
Benchmark & Method & nano & DeepSeek & Luna \\
\midrule
OfficeQA & \gepa{} & -- & $-0.6$ & $+1.4$ \\
 & \sol{} & -- & $+0.9$ & $+0.3$ \\
SpreadsheetBench & \gepa{} & -- & $\mathbf{+1.9}$ & $+0.6$ \\
 & \sol{} & -- & $\mathbf{+5.8}$ & $+0.0$ \\
HotpotQA & \gepa{} & $\mathbf{+8.6}$ & $\mathbf{+8.9}$ & $\mathbf{+3.5}$ \\
 & \sol{} & $-0.8$ & $\mathbf{+2.2}$ & $+0.0$ \\
LiveMathematicianBench & \gepa{} & $+3.1$ & $-4.3$ & $-3.5$ \\
 & \sol{} & $+1.2$ & $-3.9$ & $-3.5$ \\
\midrule
ALFWorld & HarnessLens & $+3.2$ & $-0.4$ & $-0.8$ \\
 & Meta-Harness & $-1.6$ & $\mathbf{+30.3}$ & $\mathbf{+16.4}$ \\
 & AHE & $-0.1$ & $\mathbf{+31.7}$ & $\mathbf{+23.6}$ \\
SpreadsheetBench & HarnessLens & $+0.5$ & $-0.3$ & $+1.6$ \\
 & Meta-Harness & $+0.6$ & $\mathbf{+6.3}$ & $+0.1$ \\
 & AHE & $-3.6$ & $\mathbf{+7.6}$ & $\mathbf{+8.8}$ \\
OfficeQA & HarnessLens & $-0.3$ & $-0.3$ & $-1.1$ \\
 & Meta-Harness & $+3.7$ & $\mathbf{+2.8}$ & $+0.3$ \\
 & AHE & $+2.8$ & $\mathbf{+3.7}$ & $+1.2$ \\
\bottomrule
\end{tabular}}
\caption{Gain (pp) of artifacts evolved under the repaired evaluation over the same-model baseline under that evaluation. Bold: 95\% confidence interval excludes zero. LiveMathematicianBench is scored on questions without the leaked option. No shortcut was found in the ALFWorld evaluation, so its harness rows use the released evaluation.}
\label{tab:audit-reevolve}
\end{table}

\subsection{Headroom After Repair}
\label{app:audit:headroom}

Repairing an evaluation also shows how much room a benchmark leaves for evolution.
Terminal-Bench, which we did not audit, is included under its released evaluation.

\paragraph{Benchmarks that no longer measure their target.}
On two benchmarks, the score can be earned without the capability the benchmark is meant to test.
\begin{itemize}[leftmargin=*,itemsep=2pt,topsep=2pt]
\item \textbf{SearchQA: closed-book knowledge instead of retrieval.}
The benchmark is meant to test answering from retrieved search snippets, but the models answer as well without them.
Under semantic grading, removing all snippets changes the score from 84.2 to 81.9 for GPT-5.4 nano, from 89.4 to 90.2 for DeepSeek-V4-Flash, and from 89.8 to 90.7 for GPT-5.6 Luna.
Snippets help only on the 126 of 1,400 questions whose snippets come from question-bank sites and contain both the clue and the answer.
The remaining errors say little about the models: of 100 residual errors, which we attributed with an LLM and spot-checked by hand, 36 stem from wrong or ambiguous gold answers, such as Jeopardy clues that lost their category, or from answers that are in fact acceptable.
The benchmark cannot even separate DeepSeek-V4-Flash from GPT-5.6 Luna ($-0.4$ pp, 95\% CI $[-1.5, 0.8]$), and the released skill adds at most 1.6 pp.
\item \textbf{LiveMathematicianBench: one option is always correct.}
The benchmark turns theorems from recent arXiv papers into five-option questions that ask for the strongest statement that can be proved.
Some questions include the option ``One of the remaining options is correct, but a stronger result can be proven.''
Across the 177 questions of the released splits, this option appears in 85 and is the correct answer in all 85; it appears in 46 of the 106 test questions.
For example, one test question is built from the theorem that every flow-admissible, cyclically 5-edge-connected, cubic signed graph has a nowhere-zero 6-flow.
The theorem itself is not among the options: one option weakens its conclusion to a $\mathbb{Z}_6$-flow, two strengthen it to cyclic 4-edge-connectivity or a 5-flow, one drops flow-admissibility, and the correct answer is the stronger-result option.
A model that always picks this option answers all 46 such test questions correctly without reading the mathematics, and the released skill's entire gain comes from these questions.
On the 60 test questions without the option, the baselines are 50--61, yet no artifact improves on them; with 60 questions the confidence interval is about $\pm$5 pp.
\end{itemize}

\paragraph{Little headroom.}
On several benchmarks the baseline, or the baseline plus one sentence, is already close to full marks once answers are graded correctly, so evolution has little left to gain.
\begin{itemize}[leftmargin=*,itemsep=2pt,topsep=2pt]
\item \textbf{DocVQA.} With semantic grading the baseline reaches 88--94, and the released skill adds nothing.
\item \textbf{AIME 2025.} DeepSeek-V4-Flash solves 96.5\% of attempts without evolution. For all three models the repaired gains stay within 2.5 pp, comparable to the confidence interval of a 30-problem test set.
\item \textbf{PUPA.} The evolved artifacts reach 96--97 for DeepSeek-V4-Flash and GPT-5.6 Luna, and one sentence added to the baseline reaches the same level.
\item \textbf{ALFWorld, for strong models only.} Gemini-3.7-flash, an additional model outside our main grid, starts at 96.3, and the best of our four text-artifact methods reaches 97.0. In the \sol{} paper~\citep{skilloptlite}, both \sol{} and its strongest competitor score 100.0 with GPT-5.4-mini, GPT-5.4, and GPT-5.5, so the benchmark can no longer rank methods for these models.
\end{itemize}

\paragraph{Headroom without an identifiable gain.}
On other benchmarks much room is left, yet evolution gains little or nothing that the evaluation can identify.
\begin{itemize}[leftmargin=*,itemsep=2pt,topsep=2pt]
\item \textbf{Terminal-Bench.} It leaves the most room, 28 pp for DeepSeek-V4-Flash and 46 for GPT-5.6 Luna, yet the best method gains only 1.0 and 3.6 pp, below the standard deviation of a single run (Section~\ref{sec:crisis:repetitions}).
\item \textbf{OfficeQA.} Under its official evaluation it leaves 35--39 pp, but neither the released artifacts nor re-evolved text-artifact methods gain anything, and re-evolved harness methods gain at most 3.7 pp (Table~\ref{tab:audit-reevolve}).
\item \textbf{LiveBench-Math.} It leaves 11--17 pp after repair, yet the repaired gains are only 0.2 to 1.3 pp.
\item \textbf{IFBench.} Its gains of 2.7--3.8 pp are of the same order as its evaluation noise (Appendix~\ref{app:audit:results}).
\end{itemize}

\paragraph{Headroom with a real gain.}
Three benchmarks keep a gain that the repair does not remove, and in each case its source can be named.
On HotpotQA, \gepa{} keeps 5.5--7.6 pp under semantic grading, and re-evolution traces the gain to retrieval: recall of supporting articles rises by about 10 pp.
On SpreadsheetBench, 5.7--6.7 pp survive LibreOffice recalculation, but the re-evolved artifacts that keep a gain do so by adapting to the recalculation engine, writing values and avoiding unsupported functions.
On HoVer, gains of 8--13 pp come from retrieval strategies.
These are the small, specific improvements whose identification Section~\ref{sec:inside} examines.

\subsection{Scope of the Audit}
\label{app:audit:scope}

The audit tests whether the largest reported gains need more than a single rule; it does not show that an evolved artifact contains nothing else of value.
Because we wrote each rule after inspecting the artifacts, the test shows that one rule is enough to reproduce a gain, not that the gain requires this rule, and HoVer shows that the test fails when a gain is spread over several strategies.
The artifacts released with \gepa{} were evolved for GPT-4.1-mini, so their gains on our models are smaller than those in the original paper; we therefore compare, on the same model, the gain under the released evaluation with the one-rule result and with the repaired gain.
Finally, a repair can be incomplete.
For re-evolution on LiveMathematicianBench, we balance the leaked option: in as many questions as it is correct in, we replace a wrong option with its text, so that it is correct only half the time where it appears.
Half is still more often than the models answer these questions correctly (about 34\% for DeepSeek-V4-Flash), so choosing the option remains profitable, and balancing only halves its benefit.
We therefore score capability only on the questions that do not contain the option.

%% file: figures/audit/audit_full_table.tex
\begin{tabular}{@{}llr rrr rr rrr@{}}
\toprule
& & & \multicolumn{3}{c}{Released evaluation} & \multicolumn{2}{c}{Baseline + one rule} & \multicolumn{3}{c}{Repaired evaluation} \\
\cmidrule(lr){4-6} \cmidrule(lr){7-8} \cmidrule(l){9-11}
Benchmark & Model & Items$\times$reps & Base & Evolved & Gain & Score & Gain & Base & Evolved & Gain \\
\midrule
SpreadsheetBench & nano & 281$\times$16 & 33.5 & 64.6 & \textbf{+31.1} & 62.5 & \textbf{+29.0} & 58.0 & 64.7 & \textbf{+6.7} \\
 & DeepSeek &  & 43.4 & 72.0 & \textbf{+28.6} & 69.1 & \textbf{+25.7} & 66.9 & 72.6 & \textbf{+5.7} \\
 & Luna &  & 35.3 & 73.7 & \textbf{+38.4} & 75.2 & \textbf{+39.9} & 69.4 & 75.3 & \textbf{+5.9} \\
\addlinespace[2pt]
AIME 2025 & nano & 30$\times$4 & 32.0 & 87.3 & \textbf{+55.3} & 86.7 & \textbf{+54.7} & 85.8 & 87.7 & \textbf{+1.8} \\
 & DeepSeek &  & 96.5 & 96.0 & $-$0.5 & 97.3 & +0.8 & 96.7 & 96.5 & $-$0.2 \\
 & Luna &  & 11.5 & 51.0 & \textbf{+39.5} & 69.3 & \textbf{+57.8} & 80.3 & 82.8 & +2.5 \\
\addlinespace[2pt]
LiveMathematicianBench & nano & 106$\times$4 & 29.0 & 56.1 & \textbf{+27.1} & 28.3 & $-$0.7 & 49.6 & 47.9 & $-$1.7 \\
 & DeepSeek &  & 38.9 & 68.4 & \textbf{+29.5} & 54.0 & \textbf{+15.1} & 54.6 & 54.6 & +0.0 \\
 & Luna &  & 37.0 & 73.8 & \textbf{+36.8} & 46.7 & \textbf{+9.7} & 60.8 & 59.6 & $-$1.2 \\
\addlinespace[2pt]
DocVQA & nano & 374$\times$4 & 61.0 & 82.0 & \textbf{+21.0} & 79.9 & \textbf{+18.9} & 88.3 & 88.1 & $-$0.2 \\
 & DeepSeek & \multicolumn{9}{c}{\emph{not run: no image input}} \\
 & Luna &  & 73.1 & 87.2 & \textbf{+14.1} & 88.6 & \textbf{+15.5} & 93.9 & 93.2 & $-$0.7 \\
\addlinespace[2pt]
HotpotQA & nano & 300$\times$4 & 40.2 & 63.6 & \textbf{+23.3} & 58.4 & \textbf{+18.2} & 77.0 & 82.8 & \textbf{+5.8} \\
 & DeepSeek &  & 57.4 & 71.0 & \textbf{+13.6} & 67.1 & \textbf{+9.7} & 78.3 & 85.9 & \textbf{+7.6} \\
 & Luna &  & 51.2 & 57.9 & \textbf{+6.7} & 63.8 & \textbf{+12.6} & 83.8 & 89.2 & \textbf{+5.5} \\
\addlinespace[2pt]
OfficeQA & nano & 148$\times$16 & 11.4 & 12.3 & +1.0 & 9.2 & \textbf{$-$2.1} & -- & -- & -- \\
 & DeepSeek &  & 55.3 & 58.5 & \textbf{+3.3} & 58.8 & \textbf{+3.5} & 64.7 & 64.9 & +0.2 \\
 & Luna &  & 35.5 & 56.6 & \textbf{+21.1} & 56.9 & \textbf{+21.4} & 61.0 & 60.9 & $-$0.1 \\
\addlinespace[2pt]
SearchQA & nano & 1400$\times$2 & 52.0 & 63.5 & \textbf{+11.5} & 60.9 & \textbf{+8.9} & 84.2 & 85.8 & \textbf{+1.6} \\
 & DeepSeek &  & 69.3 & 73.7 & \textbf{+4.4} & 71.3 & \textbf{+2.0} & 89.4 & 90.4 & \textbf{+1.0} \\
 & Luna &  & 63.9 & 73.0 & \textbf{+9.1} & 70.2 & \textbf{+6.4} & 89.8 & 89.6 & $-$0.1 \\
\addlinespace[2pt]
LiveBench-Math & nano & 126$\times$4 & 73.8 & 74.7 & +0.9 & -- & -- & 82.7 & 83.2 & +0.5 \\
 & DeepSeek &  & 80.5 & 81.8 & +1.3 & -- & -- & 88.6 & 90.0 & +1.3 \\
 & Luna &  & 72.9 & 76.8 & \textbf{+3.9} & -- & -- & 85.2 & 85.4 & +0.2 \\
\addlinespace[2pt]
\midrule
PUPA & nano & 221$\times$4 & 73.6 & 91.9 & \textbf{+18.3} & 92.9 & \textbf{+19.3} & -- & -- & -- \\
 & DeepSeek &  & 86.8 & 97.2 & \textbf{+10.4} & 96.2 & \textbf{+9.4} & -- & -- & -- \\
 & Luna &  & 85.7 & 96.4 & \textbf{+10.7} & 97.3 & \textbf{+11.6} & -- & -- & -- \\
\addlinespace[2pt]
ALFWorld & nano & 67$\times$16 & 49.1 & 60.6 & \textbf{+11.5} & 63.4 & \textbf{+14.4} & -- & -- & -- \\
 & DeepSeek &  & 65.0 & 76.5 & \textbf{+11.5} & 68.4 & \textbf{+3.4} & -- & -- & -- \\
 & Luna &  & 68.2 & 79.1 & \textbf{+10.9} & 76.2 & \textbf{+8.0} & -- & -- & -- \\
\addlinespace[2pt]
HoVer & nano & 300$\times$4 & 50.3 & 58.2 & \textbf{+7.8} & 53.7 & \textbf{+3.3} & -- & -- & -- \\
 & DeepSeek &  & 42.9 & 56.2 & \textbf{+13.2} & 55.0 & \textbf{+12.1} & -- & -- & -- \\
 & Luna &  & 51.2 & 60.7 & \textbf{+9.4} & 55.1 & +3.8 & -- & -- & -- \\
\addlinespace[2pt]
IFBench & nano & 294$\times$4 & 70.3 & 74.1 & \textbf{+3.8} & -- & -- & -- & -- & -- \\
 & DeepSeek &  & 77.2 & 80.4 & \textbf{+3.1} & -- & -- & -- & -- & -- \\
 & Luna &  & 66.1 & 68.8 & +2.7 & -- & -- & -- & -- & -- \\
\bottomrule
\end{tabular}

%% file: sections/app_pool.tex
\subsection{Candidate Pools, Timeouts, and Evaluation Protocols}
\label{app:pool}

Sections~\ref{sec:proposals} and~\ref{sec:selection} evaluate two fixed pools of $n{=}100$ skills designed for the two stages of candidate-level evaluation: the \textbf{Proposal Pool}, which tests out-of-sample skill transfer from 10 training tasks to 10 disjoint test tasks, and the \textbf{Selection Pool}, which benchmarks selectors and trajectory judges across 11 difficulty-stratified tasks.

\paragraph{Proposal Pool (Section~\ref{sec:proposals}).}
To evaluate proposal quality without selection in the loop, an offline \sol{}-based proposer using DeepSeek-V4-Flash generates $n{=}100$ standalone skills from 78 execution trajectories across 10 Terminal-Bench v2.1 training tasks (kv-store-grpc, pypi-server, sqlite-db-truncate, path-tracing, filter-js-from-html, torch-tensor-parallelism, dna-insert, mcmc-sampling-stan, bn-fit-modify, tune-mjcf) following the protocol below.
Each skill is then evaluated 16 times on each of 10 disjoint Terminal-Bench v2.1 test tasks (configure-git-webserver, break-filter-js-from-html, query-optimize, dna-assembly, rstan-to-pystan, vulnerable-secret, mailman, db-wal-recovery, torch-pipeline-parallelism, path-tracing-reverse), totaling $100 \times 10 \times 16 = 16{,}000$ solver rollouts with binary whole-task rewards under native task verifiers.

\begin{figure*}[t]
\centering
\includegraphics[width=0.98\textwidth]{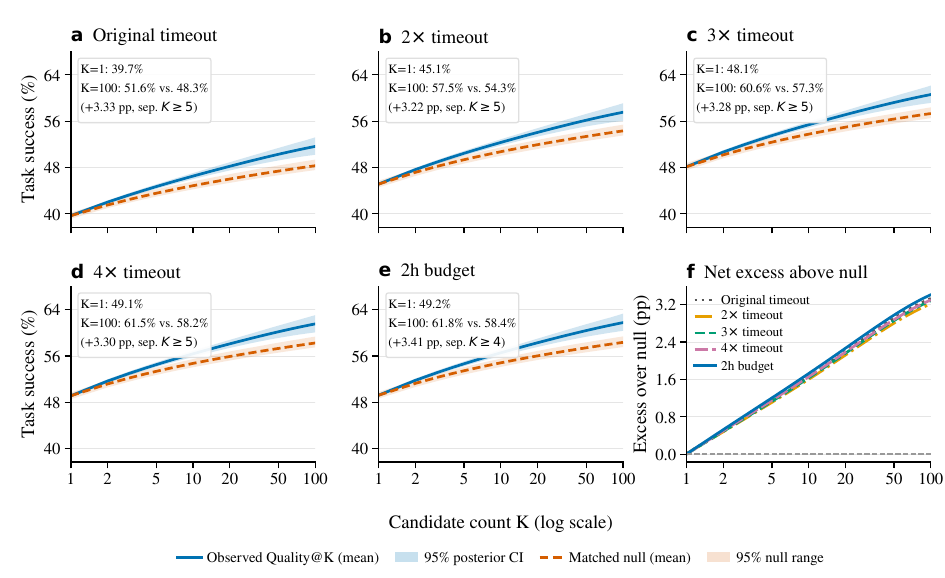}
\caption{Effect of agent timeout cutoff on Quality@$K$ vs.\ the matched equal-quality null on the Proposal Pool (10 disjoint Terminal-Bench v2.1 test tasks, $n{=}100$ skills, $\kappa_{\max}{=}100$ empirical-Bayes estimator). Panels (a)--(e) evaluate the same 16,000 trajectories under progressively relaxed cutoffs: original per-task timeout ($900\text{--}1800$\,s), $2\times$, $3\times$, $4\times$, and the full 2-hour budget. Panel (f) overlays the net excess above the matched null across all five cutoffs.}
\label{fig:qualityk-timeouts}
\end{figure*}

\begin{figure*}[t]
\centering
\includegraphics[width=0.98\textwidth]{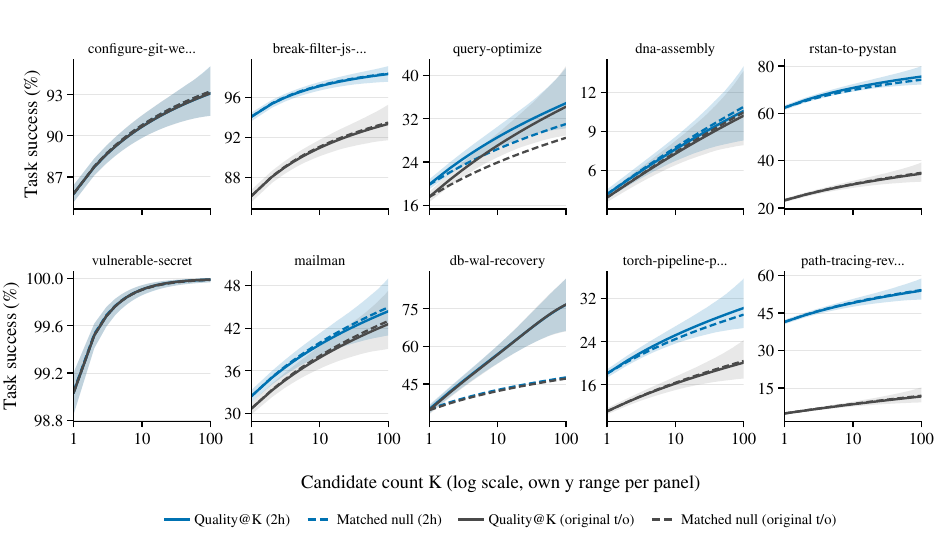}
\caption{Per-task Quality@$K$ (solid) and matched equal-quality null (dashed) across all 10 disjoint Terminal-Bench v2.1 test tasks of the Proposal Pool under the 2-hour budget (blue) and original per-task timeout (charcoal), using $\kappa_{\max}{=}100$ empirical Bayes; each panel uses its own vertical scale.}
\label{fig:qualityk-test10-pertask}
\end{figure*}

\paragraph{Original timeout vs.\ extended execution budgets.}
All 16,000 rollouts are executed with a 2-hour (7,200\,s) agent budget.
Because Terminal-Bench tasks normally enforce shorter native timeouts (900\,s, 1,200\,s, or 1,800\,s per task), we also score five timeout cutoffs on the exact same set of finished trajectories by marking a rollout as failed whenever agent elapsed time exceeds the cutoff: the task's \emph{original} timeout, $2\times$, $3\times$, $4\times$, and the full \emph{2h} budget.
Figure~\ref{fig:qualityk-timeouts} shows Quality@$K$ and its matched equal-quality null across all five cutoffs under our primary empirical-Bayes estimator ($\kappa_{\max}{=}100$; Appendix~\ref{app:qualityatk}):
the pooled $K{=}1$ success rate rises from $39.69\%$ under original timeouts through $45.15\%$ ($2\times$), $48.14\%$ ($3\times$), and $49.13\%$ ($4\times$) to $49.23\%$ at 2\,h, while the $K{=}100$ excess above the matched null remains cleanly positive across every cutoff at $+3.33$, $+3.22$, $+3.28$, $+3.30$, and $+3.41$ pp (with non-overlapping 95\% bands from $K\ge 5$ under original--$4\times$ timeouts and from $K\ge 4$ at 2\,h).
Figure~\ref{fig:qualityk-test10-pertask} displays all 10 individual test tasks under both original and 2\,h timeouts, showing that tasks whose multi-step compilation or rendering exceeds native limits (such as \texttt{path-tracing-reverse}, rising from $4.8\%$ at original timeout to $41.4\%$ at 2\,h, and \texttt{rstan-to-pystan}, rising from $23.2\%$ to $62.4\%$) explain most of the level shift between cutoffs, while near-homogeneous tasks track their matched equal-quality nulls closely within posterior bands.

\paragraph{Selection Pool (Section~\ref{sec:selection}).}
To benchmark selection operators and LLM trajectory judges across diverse candidates and task difficulty levels, Section~\ref{sec:selection} evaluates the Selection Pool: $n{=}100$ skills generated from execution trajectories across six Terminal-Bench v2.1 tasks and three DeepSWE tasks, rolled out on 11 Terminal-Bench tasks spanning five baseline difficulty levels (six source tasks and five held out from skill generation; listed in Appendix~\ref{app:setup}).
Every artifact--task pair receives 8 rollouts, extended to 15 for the ten highest-scoring artifacts per task, and the unevolved baseline receives 12 rollouts per task.
Its per-task selection curves are reported in Appendix~\ref{app:grouped-curves}.

\subsubsection{Proposer Research Protocol}
\label{app:proposer-prompt}

To generate $n{=}100$ distinct offline candidate skills from the training trajectories without an outer selection loop, the proposer launches parallel research sub-agents that inspect failure mechanisms on the training corpus and verify candidate fixes with small local probes.
Below is the research and skill-authoring prompt used to synthesize the $n{=}100$ standalone skills in the Proposal Pool.

\begin{tcolorbox}[breakable, colback=gray!3, colframe=gray!40,
  boxrule=0.4pt, arc=1.2mm, left=2.5mm, right=2.5mm, top=2mm, bottom=2mm,
  fontupper=\scriptsize]
\textbf{Research \& Skill Authoring Guidance}\par\smallskip
Your goal is to investigate the training corpus in \texttt{workspace/.skillopt/corpus/}, uncover the concrete mechanisms behind task successes and failures, verify them, and translate them into one hundred complete, standalone actionable skills at \texttt{workspace/artifacts/001/SKILL.md} through \texttt{workspace/artifacts/100/SKILL.md} (each with its own \texttt{report.md}), backed by an auditable evidence report in \texttt{workspace/report.md}.\par\smallskip
There is no fixed reading quota, edit limit, or rigid output template. Read, search, write scripts, and run small executable probes in \texttt{workspace/} as much as the evidence warrants.\par\medskip

\textbf{1.\ Where to look: find the ``moment of wrong commitment''}\par
Because task outcomes in the training pool are unbalanced, do not simply group failed runs by task name. Study \textbf{both} tasks with failures (comparing failed runs against passing runs on the same task to see the full workflow past early setup/install crashes) \textbf{and} pass-only or fail-only tasks (extracting how passing runs diagnose and recover from mid-run obstacles, or using local probes in \texttt{workspace/} when no passing run exists). Focus on the critical decision points where an agent locks into a flawed path:
\begin{itemize}[leftmargin=4mm, itemsep=1.5pt, topsep=2pt]
\item \textbf{State and deliverable lifecycle:} Check whether agents mutate raw inputs in-place before making a safe backup, spend the entire budget in scratch experiments without maintaining a valid working deliverable at the target path, or accidentally undo working state and stop required services during final cleanup.
\item \textbf{Unverified references \& circular self-tests:} Agents often believe a flawed self-test---or worse, \textbf{modify a correct implementation to match an inaccurate reference}. Check whether the agent trusted a library helper or reference script as an oracle without first verifying that its implicit behavior matches the task contract, or wrote a checker that simply reused the generator's own assumptions and offsets.
\item \textbf{Premature scaling without small probes:} Check where agents launch expensive full-scale runs before using a tiny run to verify interfaces, output structure, and how runtime, memory, or output sizes scale.
\item \textbf{Stopping at surface gains:} In optimization or reconstruction tasks, check where agents stop after an initial improvement over a naive baseline or a rough approximation, versus how successful runs use concrete diagnostics (profiles, execution plans, structural parsers, or quantitative residuals) to locate and remove the remaining bottleneck.
\end{itemize}\par\smallskip

\textbf{2.\ Evidence discipline: verify mechanisms, do not just tell stories}\par
A failed run and a plausible narrative do not establish a cause, and a passing run's own explanation or code comment is only a hypothesis.
\begin{itemize}[leftmargin=4mm, itemsep=1.5pt, topsep=2pt]
\item \textbf{Inspect middle logs (beware output truncation):} Long command outputs in your environment are truncated in the middle. Never assume a broad dump showed the full trace. Before attributing a failure to downstream visibility or configuration, inspect the middle of the transcript to verify whether upstream steps (such as compilation, linking, or installation) actually succeeded.
\item \textbf{Account for exact assertions:} Repeated runs of a task are not independent tasks, and failing the same test function does not mean failing for the same reason. Check the actual verifier output per run to distinguish early format or missing-file failures from downstream numerical or behavioral failures.
\item \textbf{Verify technical claims with small probes:} Agents in passing runs sometimes rely on lucky approximations or state inaccurate formulas, byte layouts, or API rules. Before encoding a specific mathematical formula, binary/database record format, parser behavior, or library rule into the skill, verify it with a small executable probe in \texttt{workspace/} or against local documentation.
\end{itemize}\par\smallskip

\textbf{3.\ Authoring \texttt{skill.md}: concrete actions, de-anchored from training literals}\par
High-level principles are research lenses for \textbf{you (the proposer)}---they are \textbf{not} finished instructions for the solver. At the same time, a skill that is over-fitted to one training problem's surface details will fail to trigger on neighboring tasks.
\begin{itemize}[leftmargin=4mm, itemsep=1.5pt, topsep=2pt]
\item \textbf{Strip training-instance literals, keep technical depth:} Never leave training-specific filenames, JSON field names, sample IDs, or narrow one-task formulas as the trigger or example in \texttt{skill.md}. Broaden the trigger to the general technical scenario (e.g., ``corrupt database/file recovery'' rather than only ``truncated file''; ``structured sequence/protocol assembly'' rather than one specific splicing formula), while keeping the concrete diagnostic commands, byte/format rules, and verification recipes sharp.
\item \textbf{Turn principles into decision-point checks:} For each workflow you write, give the solver a complete, executable playbook:
  \begin{enumerate}[leftmargin=4mm, itemsep=1pt, topsep=1pt]
  \item \textbf{When it triggers} and \textbf{what to protect or establish first} (e.g., full directory backup before first tool contact; tiny-budget schema and memory-growth check; saving an initial compilable baseline deliverable).
  \item \textbf{How to calibrate the reference check:} Teach the solver to verify any third-party reference API on a tiny 2--3 element hand-checked input before trusting it as an oracle, and to validate structured/composite outputs with an independent parser and end-to-end simulation (without reusing builder index offsets or dropping required protocol prefixes to guess grader behavior).
  \item \textbf{How to diagnose the next bottleneck:} Provide concrete diagnostic loops (inspecting execution plans/profiles, reducing candidate sets before heavy computation, driving reconstruction via quantitative residuals, and checking multi-component service chains from a fresh client).
  \end{enumerate}
\end{itemize}\par\smallskip

\textbf{4.\ Outputs and honest boundaries}\par
All training evidence is mounted read-only at \texttt{workspace/.skillopt/corpus/} (see \texttt{corpus/README.md} for optional helpers). Use \texttt{workspace/} for your own scripts, notes, and executable probes. Do not retrieve external benchmark solutions or test-set material.
\begin{itemize}[leftmargin=4mm, itemsep=1.5pt, topsep=2pt]
\item Save exactly one hundred complete standalone skills at the artifact paths above, each free of training task names, sample IDs, and task-local literal constants.
\item Save a concise, auditable report in \texttt{workspace/report.md} tracing the chain from \textbf{what you read $\to$ what you verified via probes $\to$ what you wrote into the skill $\to$ what counterexamples or open gaps remain}, citing exact message IDs, line ranges, and verifier assertions.
\end{itemize}
\end{tcolorbox}

%% file: sections/app_qualityatk.tex
\subsection{Posterior Estimation of Quality@$K$}
\label{app:qualityatk}

Each artifact's per-task quality is a latent success probability observed through a handful of binary evaluations (Appendix~\ref{app:pool}).
Raw success counts are too noisy to rank individual artifacts reliably, so we regularize with an empirical-Bayes hierarchical model and define two complementary estimators.

\paragraph{Hierarchical model.}
Let $n$ denote the pool size, $s_i$ and $n_i$ the number of successes and total evaluations for artifact $i$ on a given task, and $\theta_i \in [0,1]$ its latent success probability.
All $n$ artifacts share a task-specific beta prior $\theta_i \sim \text{Beta}(a,b)$, where the mean $\mu = a/(a{+}b)$ and concentration $\kappa = a{+}b$ are estimated by cross-artifact method-of-moments from the pooled outcomes.
In our primary estimator, $\kappa$ is bounded in $[0.25, \kappa_{\max}]$ with $\kappa_{\max}{=}100$ (already more than $6\times$ the per-skill rollout budget $m{=}16$): whenever cross-artifact sample variance does not exceed pure binomial sampling noise, $\hat\kappa$ saturates at $\kappa_{\max}$ rather than collapsing deterministically to a zero-width point mass.
For any finite $\kappa$, the posterior for each artifact is $\theta_i \mid s_i, n_i \sim \text{Beta}(a{+}s_i,\; b{+}n_i{-}s_i)$, with posterior mean $\bar\theta_i = (a{+}s_i)/(a{+}b{+}n_i)$.
We evaluate this same $\kappa_{\max}{=}100$ moment estimator identically on both observed pools and simulated equal-quality null pools, and test sensitivity across $\kappa_{\max} \in \{30, 100, 300, \infty\}$ below.

\paragraph{Posterior Oracle.}
Quality@$K$ asks: if we draw $K$ artifacts uniformly from the pool, how good is the best one?
Let $I_K$ denote a uniformly random size-$K$ subset of the $n$ artifacts (baseline excluded).
The \emph{posterior Oracle} integrates this quantity over the full posterior:
\begin{equation}
\label{eq:quality-oracle}
\widehat{\text{Quality@}K}_{\text{Oracle}}
= 100\,\mathbb{E}_{\boldsymbol\theta\mid\mathcal{D}}
  \!\left[\,\mathbb{E}_{I_K}\!\left[\max_{i\in I_K}\theta_i\right]\right].
\end{equation}
We approximate the outer expectation with 8{,}000 posterior draws (256 draws per simulation inside each null dataset).
Within each draw, the inner expectation is exact: sorting the $n$ sampled qualities in ascending order, the artifact at rank $r$ contributes with weight $\binom{r-1}{K-1}/\binom{n}{K}$ for $r \geq K$.
Curves show draw means; shading denotes pointwise 95\% credible intervals conditional on the fitted hyperparameters.

\begin{figure*}[t]
\centering
\includegraphics[width=0.95\textwidth]{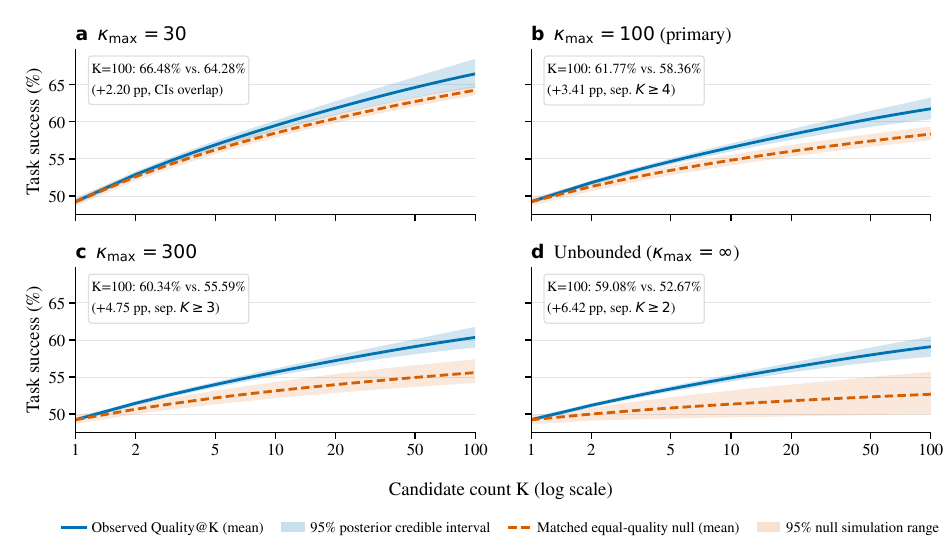}
\caption{Sensitivity of macro-averaged Quality@$K$ vs.\ matched equal-quality null under the 2-hour budget to the upper concentration cap $\kappa_{\max} \in \{30, 100, 300, \infty\}$ of the empirical-Bayes Beta prior. A tight cap ($\kappa_{\max}{=}30$) inflates dispersion in homogeneous null pools, whereas $\kappa_{\max}{=}100$ (primary) through unbounded ($\kappa_{\max}{=}\infty$) behave consistently and cleanly separate observed from null.}
\label{fig:qualityk-kappa-sensitivity}
\end{figure*}

\paragraph{Sensitivity to $\kappa_{\max}$.}
Figure~\ref{fig:qualityk-kappa-sensitivity} examines sensitivity of the 2-hour Quality@$K$ curve on the Proposal Pool across $\kappa_{\max} \in \{30, 100, 300, \infty\}$.
An overly tight cap such as $\kappa_{\max}=30$ forces substantial prior dispersion even when an equal-quality null exhibits no overdispersion, inflating the 2\,h matched null to $64.28\%$ at $K{=}100$ so the $+2.20$ pp excess lies within overlapping bands.
At $\kappa_{\max}=100$, the 2\,h curve reaches $61.77\%$ vs.\ $58.36\%$ null ($+3.41$ pp, separating cleanly for $K \ge 4$), and relaxing $\kappa_{\max}$ further to $300$ or $\infty$ widens the excess smoothly to $+4.75$ pp ($K \ge 3$) and $+6.42$ pp ($K \ge 2$).
Importantly, finite regularization at $\kappa_{\max}=100$ avoids a subtle point-mass degeneracy that arises under $\kappa_{\max}=\infty$: on near-homogeneous tasks whose sample variance sits just below the binomial floor (e.g., \texttt{dna-assembly} or \texttt{mailman}), $\kappa_{\max}=\infty$ collapses the observed prior to $\delta_\mu$ (a flat curve) whereas roughly half of pure-binomial null simulations happen to draw positive finite-sample overdispersion and finite $\hat\kappa$, nudging the simulated null mean slightly above the collapsed observed curve in per-task views.
Bounding $\kappa_{\max}=100$ keeps null-like tasks centered inside their matched null bands while every $\kappa_{\max} \ge 100$ confirms the same positive headroom conclusion.

\paragraph{Equal-quality null.}
Both estimators rise with $K$ even when all artifacts are equally good, because selecting the luckiest among more candidates yields a higher expected score.
To calibrate this statistical selection effect, we run 1{,}000 null simulations per task: all $n$ candidates are generated i.i.d.\ at the task-level pooled pass rate, and the exact same estimator (prior fitting and posterior Quality@$K$) is refit on each simulated pool.
Each estimator has its own null curve (dashed in Figures~\ref{fig:quality-selection}a, \ref{fig:qualityk-timeouts}, and \ref{fig:qualityk-kappa-sensitivity}).
Null intervals describe central 95\% simulation variability across equal-quality datasets rather than posterior uncertainty.

\paragraph{Scope.}
Quality@$K$ characterizes the finite pool of $n{=}100$ candidates, not the theoretical task optimum, and cross-task averages allow each task to select a different skill.
For the Selection Pool (Section~\ref{sec:selection}), quality ground truth uses a held-out rollout reference independent of each selector's observed rollouts (Appendix~\ref{app:selection}).

%% file: sections/app_selection.tex
\subsection{Selection Operators and Evaluation Metrics}
\label{app:selection}

\paragraph{Independent-reference protocol.}
Each selector sees only the first rollout of each artifact (the native \sol{} and \gepa{} rows, the first two).
Artifact quality is evaluated against a held-out reference consisting of the remaining rollouts that the selector has not observed; the baseline reference likewise excludes its first two rollouts.
This separation ensures that the quality ground truth used to assess selectors is independent of the information available to them.
Artifact quality posteriors are fitted to this held-out evidence as described in Appendix~\ref{app:qualityatk}.
All selectors share the same posterior draws, so selected quality is never larger than the oracle for the same draw, and all methods coincide at $K{=}1$.

\paragraph{Native parent-selection ablation.}
The \sol{} and \gepa{} rows (marked \dag{} in Table~\ref{tab:selection}) apply each method's own accept/reject logic to the first two rollout outcomes.
\sol{} replaces the current artifact only on strict improvement of the two-rollout mean; \gepa{} applies its frontier-pruning routine and samples a parent in proportion to surviving frequency.
These rows use the same held-out quality reference as the other selectors.

\paragraph{Trajectory judges.}
DeepSeek-V4-Flash serves as the judge model on the 1{,}111 artifact--task pairs (including baselines) in the Terminal-Bench subset.
Each call receives the task description, initial skill context, and the solver trajectory from the first rollout, without outcome labels.
Pointwise scoring produces one score per pair (1{,}111 calls).
Pairwise scoring evaluates 1{,}100 fixed pairs in both presentation orders (2{,}200 calls).
The LLM judge (ACE) condition adapts the ACE Reflector's no-ground-truth template (error identification, root cause, corrective approach, and helpful/harmful/neutral tags), adding an explicit success-probability score.
No ACE training loop or verifier feedback is used.

\paragraph{Reported metrics.}
\emph{Pairwise accuracy}: posterior probability that the selector's chosen artifact has higher latent quality, with ties broken uniformly.
\emph{False acceptance}: among accepted baseline replacements, the posterior fraction with quality below baseline.
\emph{Improvement and regression rates}: posterior probability that the selected artifact's quality exceeds or falls below baseline.
Spearman correlations are within-task; tasks with constant selector scores are excluded.

\paragraph{Can Higher-Resolution Selection Improve Evolution?}
\label{app:selection:higher}
Our findings raise a natural question: if we reduce evaluation noise during evolution by averaging many validation rollouts at every selection step, does evolution accumulate larger gains?
To test this directly, we run \sol{} for up to eight rounds across 16 independent evolution runs, averaging 16 validation rollouts per candidate instead of one (and evaluating final artifacts with 32 independent test rollouts per task).
Averaging 16 validation rollouts raises the mean final test score across runs from 61.4\% to 64.4\%, and raises the best evolved artifact's test score from 64.8\% to 72.7\%.

Inspecting the trajectories shows why: under single-rollout validation, an ordinary candidate occasionally draws an unusually high validation score by luck (e.g., 80\%) and becomes the incumbent; subsequent candidates cannot beat this inflated threshold, stalling evolution within four rounds.
Averaging 16 rollouts suppresses these false peaks so that accepted updates reflect real improvements, allowing the best evolved artifact to keep improving as more runs are launched.
At the same time, $16\times$ rollout validation is computationally expensive and makes greedy acceptance more conservative once a strong incumbent is found, motivating the cheaper, finer-grained behavioral diagnostics explored in Section~\ref{sec:forward}.

%% file: sections/12b_appendix_grouped_curves.tex
\subsection{Per-Task Selection Curves}
\label{app:grouped-curves}

\begin{figure*}[p]
\centering
\includegraphics[width=0.96\textwidth]{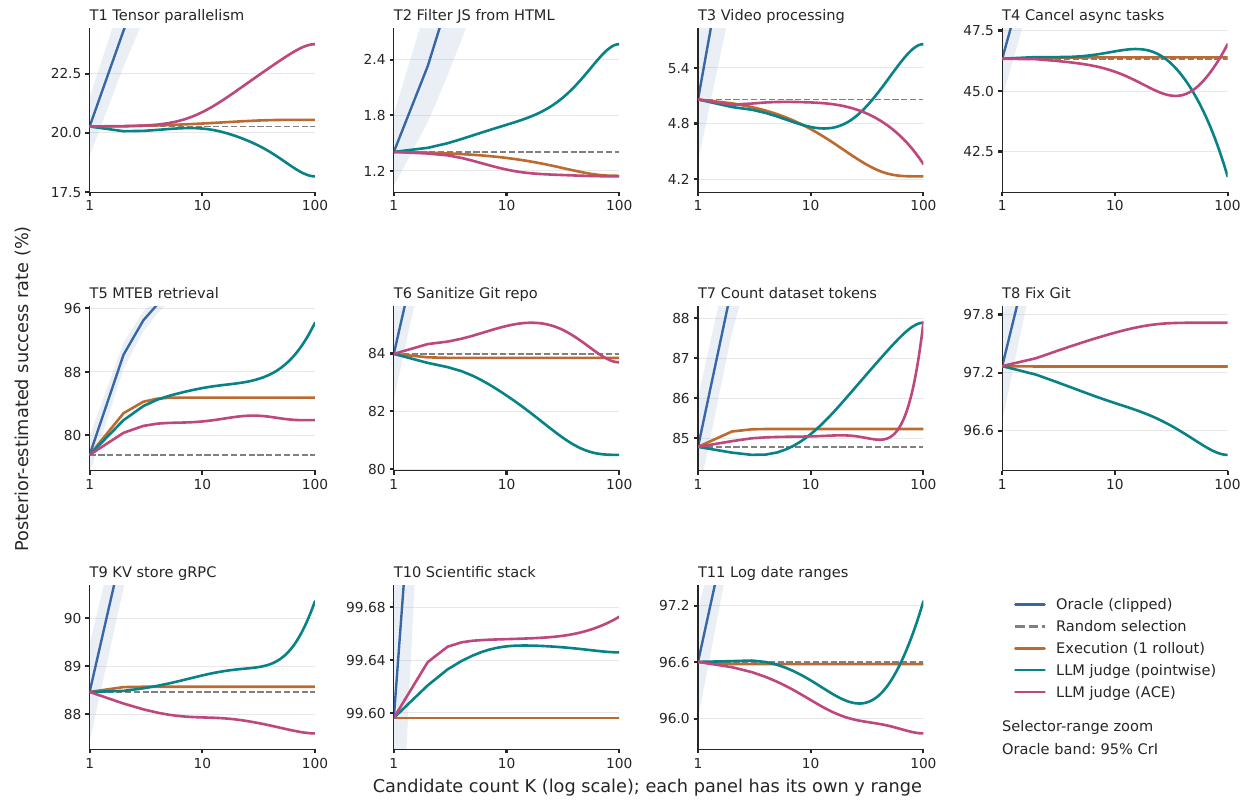}
\caption{Per-task selection curves underlying Figure~\ref{fig:quality-selection}b across the 11 tasks of the Selection Pool. Each panel shows selected-artifact success probability under the held-out quality reference (Appendix~\ref{app:selection}); the oracle and its credible band are clipped to the selector range.}
\label{fig:selection-per-task}
\end{figure*}

Figure~\ref{fig:selection-per-task} decomposes the aggregate selection curve of Figure~\ref{fig:quality-selection}b across each of the 11 Terminal-Bench v2.1 tasks in the Selection Pool (six source tasks T2, T5, T7, T9--T11 and five held-out tasks T1, T3--T4, T6, T8).
Each panel plots selected-artifact success probability as a function of the candidate budget $K$ under the held-out rollout reference (Appendix~\ref{app:selection}).

\paragraph{Equivalent budget.}
For each task and selector, the equivalent budget $K_{\mathrm{eq}}$ is the smallest integer $K$ at which the oracle's posterior mean quality reaches or exceeds the selector's quality at $K{=}100$.
It measures how many candidates an oracle selector would need to match a given selector after the latter has examined all $n{=}100$ candidates.

%% file: sections/app_accumulation.tex
\section{Trajectory Re-evaluation}
\label{app:accumulation}

We recover eight independent evolution runs per method and model for \sol{} and \gepa{} with DeepSeek-V4-Flash and GPT-5.6 Luna.
The 119 distinct within-run artifact versions are independently re-evaluated on the same 18 Terminal-Bench validation tasks, targeting eight rollouts per artifact--task pair (versus one during evolution).

\paragraph{Variance estimation.}
We estimate $\tau_t^2$ as the sample variance of the eight task-averaged re-evaluation means, subtracting the mean estimated variance of their evaluation noise.
For task success estimate $\hat p_{rjt}$ from $n_{rjt}$ available repeats, the correction for run $r$ is $18^{-2}\sum_j\hat p_{rjt}(1-\hat p_{rjt})/(n_{rjt}-1)$ on the probability scale.
Unlike Appendix~\ref{app:estimators}, which reports $\hat{\tau}$ itself and therefore sets negative estimates to zero, we retain negative corrected estimates here, because we compare $\hat{\tau}_t^2$ across rounds and clipping would bias the small early-round estimates upward.

%% file: sections/app_wayout.tex
\section{Low-Variance Process Diagnostics}
\label{app:forward}
\label{app:forward:variance}

\paragraph{Setup.}
We reuse the trajectories of the variance experiments (Section~\ref{sec:crisis}) on DeepSeek-V4-Flash.
The terminal reward is the benchmark's pass/fail verdict.
The process metric is the fraction of check commands along the trajectory whose failure diagnostics are not masked, where a check command verifies intermediate state, such as running a test after an edit, and its diagnostics are masked if the agent cannot tell from the output that something failed.
Both quantities are computed per trajectory, so their variances are directly comparable.

\paragraph{Estimation.}
For each configuration, we estimate the evaluation noise of each quantity with the within-artifact estimator of Appendix~\ref{app:estimators}, applied to the repeated evaluations of the same artifact.

\begin{table}[t]
\centering
\footnotesize
\setlength{\tabcolsep}{3pt}
\begin{tabularx}{\columnwidth}{@{}Xccr@{}}
\toprule
Method & $\sigma^2_{\text{reward}}$ & $\sigma^2_{\text{behavior}}$ & Ratio \\
\midrule
Baseline       & 0.073 & 0.0013 & 57$\times$ \\
\gepa{} \citep{agrawal2025gepareflectivepromptevolution}       & 0.074 & 0.0026 & 28$\times$ \\
HarnessLens \citep{xu2026verifysmarterevolvefurther}   & 0.056 & 0.0044 & 13$\times$ \\
Meta-Harness \citep{lee2026metaharnessendtoendoptimizationmodel}   & 0.084 & 0.0015 & 57$\times$ \\
\rbank{} \citep{ouyang2026reasoningbank}      & 0.075 & 0.0021 & 35$\times$ \\
AHE \citep{lin2026ahe} & 0.095 & 0.0022 & 44$\times$ \\
\sol{} \citep{skilloptlite}        & 0.071 & 0.0013 & 56$\times$ \\
\tts{} \citep{trace2skill2026}        & 0.077 & 0.0021 & 37$\times$ \\
\bottomrule
\end{tabularx}
\caption{Variance of the evaluation noise: terminal pass/fail vs.\ a process metric (fraction of check commands without masked failure diagnostics), DeepSeek-V4-Flash.}
\label{tab:process-variance}
\end{table}

For the baseline and all seven methods, the evaluation noise of the process metric is 13 to 57 times smaller than that of the terminal reward.

\paragraph{Behavioral shifts visible only to the process metric.}
The post-edit verification rate is the fraction of file modifications immediately followed by a test execution or sanity check.
On DeepSeek-V4-Flash, \tts{} raises this rate by $+8.73$ pp over baseline ($z = 2.17$ against the evaluation noise of the difference).
On GPT-5.6 Luna, AHE lowers it by $17.98$ pp (variance of the difference $0.0018$, $z = -4.19$).
Neither shift is detectable in the terminal success rate of the same runs.

\paragraph{Reading.}
A process metric is a diagnostic, not a selector: a candidate that passes more checks need not solve more tasks, and a selector driven by such checks would face its own resolution question.
Its value is that it makes behavioral change visible at resolutions where the terminal verdict is blind.

%% file: sections/app_shared_artifacts.tex
\newtcolorbox{artifactbox}{breakable, colback=gray!4, colframe=gray!45,
  boxrule=0.4pt, arc=1.2mm, left=2mm, right=2mm, top=1.5mm, bottom=1.5mm,
  fontupper=\footnotesize}
\newcommand{\pitfall}[4]{\noindent\textbf{Agent pitfall:} #1\par
  \textbf{Trigger:} #2\par
  \textbf{Concrete correction:} #3\par
  \textbf{Tiny verification:} #4\par\smallskip}

\subsection{Shareable Artifacts in the Candidate Pool}
\label{app:shared-artifacts}

Section~\ref{sec:proposals} shows that candidate skills vary substantially even when a single skill is shared across tasks.
To illustrate what effective shared and task-specific skills look like in practice, this appendix prints three concrete skills from the Selection Pool (Appendix~\ref{app:pool}): seed-027 and seed-099, tied for the best average gain across all 11 Terminal-Bench v2.1 tasks (+11.4 pp over baseline), and seed-058, which achieves a perfect score on mteb-retrieve in every rollout yet adds little when shared globally.

\subsubsection{Artifact seed-027 (best average gain, +11.4 pp)}
\label{app:artifact-027}

\begin{artifactbox}
\textbf{Software Engineering Skill}
\begin{itemize}[leftmargin=4mm, itemsep=0pt, topsep=1pt]
\item Read the task and relevant source before changing code.
\item Preserve existing behavior while implementing the requested public API.
\item Make a coherent implementation across affected modules.
\item Verify normal behavior and edge cases with focused tests.
\item Treat every precise validation, parsing, error, export, and format rule in the prompt as an explicit assertion.
\item Run relevant existing tests and address regressions.
\end{itemize}

\medskip

\pitfall{Substituting an idiomatic library name, conventional class name, or a generalized ``reasonable'' signature for the exact exported symbol, function signature, return shape, or error-message string that the task explicitly names.}
{A prompt that gives a verbatim public identifier (class/function/entrypoint) or a literal error/status text that a hidden caller or assertion will match, while the agent invents a library-flavored alternative such as a generated servicer class name.}
{Before coding, extract every verbatim identifier, signature, empty/zero-value return rule, and error fragment from the task; export and reference those exact strings on the public surface and error paths, and add a focused check asserting each named symbol and message exists exactly as spelled.}
{For a spec that exports \texttt{Mean(xs []int) (float64, error)} and must return \texttt{(0, nil)} on an empty slice, expose exactly \texttt{Mean} (not \texttt{computeMean}) and return the documented \texttt{(0, nil)} instead of \texttt{(-1, err)}, then confirm with one empty-input call.}

\pitfall{Improvising tokenization or embedding calls by hand (custom separators, toggling special-token flags, or a single generic task/prompt name applied to both query and corpus) instead of delegating to the named model/package and its documented encoding conventions.}
{A data task that names a specific tokenizer, embedding model, or package and requires an exact numeric count or ranking derived from that model's own encoding behavior.}
{Load the model only through the named package/entrypoint and let it both tokenize and encode every input; when the wrapper distinguishes query vs.\ passage or exposes special-token/clip parameters, set those explicitly per side rather than passing one default, and recompute the count or ranking a second time to confirm it is stable.}
{Two short strings joined by \texttt{"\textbackslash n"} and \texttt{" "} should tokenize to the same total under the named tokenizer's documented defaults; assert that changing the separator or the query/passage prompt flag alters the result, then keep the documented default so the final number matches the scorer's.}
\end{artifactbox}

\subsubsection{Artifact seed-099 (tied best, +11.4 pp)}
\label{app:artifact-099}

\begin{artifactbox}
\textbf{Software Engineering Skill}
\begin{itemize}[leftmargin=4mm, itemsep=0pt, topsep=1pt]
\item Read the task and relevant source before changing code.
\item Preserve existing behavior while implementing the requested public API.
\item Make a coherent implementation across affected modules.
\item Verify normal behavior and edge cases with focused tests.
\item Treat every precise validation, parsing, error, export, and format rule in the prompt as an explicit assertion.
\item Run relevant existing tests and address regressions.
\end{itemize}

\medskip

\pitfall{Assuming an exact, machine-checked numeric result produced by a third-party text pipeline (a tokenizer count, an embedding similarity rank, a retrieved document) is invariant to how the input text is assembled, so you join source fields with an arbitrary separator or encode both the query and the reference set with the same default prompt setting.}
{A script must output one exact number derived from feeding text to an encoder, and it builds that text by concatenating several source fields, or it calls a single encode pass for both a probe query and a corpus of documents.}
{Replicate the source data-shape exactly: join fields with the same boundary character used in the corpus (a newline, not a space, can change the token count by one or more), and when the encoder wrapper is instruction-aware, call it once with the query prompt type for the probe and once with the passage prompt type for the documents so each side receives its real instruction prefix.}
{\texttt{len(encode("a\textbackslash nb"))} is not required to equal \texttt{len(encode("a b"))}, and the same text embedded under a query prompt differs from a passage prompt, so picking the wrong role can move a different document into 5th place.}

\pitfall{Renaming or re-wrapping the exact public symbol (class, function, or variable) that the prompt explicitly names, substituting a more idiomatically descriptive name, on the assumption that the checker will call whatever the agent chose to define.}
{The task text names an importable entrypoint in fixed wording, for example ``in a class called X'', ``expose \texttt{BeginRepl(args, version)}'', or a named API variable, and the agent instead ships a differently-named implementation.}
{Declare the prompt-named symbol verbatim (exact name and signature) as a thin exported shim and delegate the real work to internal helpers; treat any renaming as a broken contract because the graded harness imports and calls the literal name, not a paraphrase.}
{If a spec requires \texttt{class Server}, the harness does \texttt{from module import Server}; a class shipped under a name like \texttt{Servicer} is untouched by that import and fails even though it runs correctly by hand.}
\end{artifactbox}

\subsubsection{Artifact seed-058 (best on mteb-retrieve)}
\label{app:artifact-058}

\begin{artifactbox}
\textbf{Software Engineering Skill}
\begin{itemize}[leftmargin=4mm, itemsep=0pt, topsep=1pt]
\item Read the task and relevant source before changing code.
\item Preserve existing behavior while implementing the requested public API.
\item Make a coherent implementation across affected modules.
\item Verify normal behavior and edge cases with focused tests.
\item Treat every precise validation, parsing, error, export, and format rule in the prompt as an explicit assertion.
\item Run relevant existing tests and address regressions.
\end{itemize}

\medskip

\pitfall{When feeding text to a subword tokenizer or a retrieval encoder from an external AI library, the agent guessed the assembled text shape (joining related fields with a space or nothing) and the special-token flag, and guessed the per-role prompt, instead of reading the documented contract.}
{A counting or retrieval result that is sensitive to how the exact text is assembled and to which prompt is applied: combining two text fields with a space versus a newline, a guessed special-token setting, or encoding the query and the documents with the same prompt while task-aware encoders distinguish them, can shift a token count or change which document ranks $N$th.}
{Check the source/package documentation for the exact text format and prompt roles before encoding.  Join independent text fields with the documented separator (BPE tokenizers merge a leading space into the following subword, so a newline avoids cross-field merging) and set the tokenizer's special-token flag deliberately.  With a retrieval encoder, encode the query with the query prompt and the documents with the passage prompt (e.g.\ a \texttt{prompt\_type}/role argument), never one shared default for both.}
{For a BPE tokenizer, tokenizing \texttt{"a" + " " + "b"} returns a different subword count than \texttt{"a" + "\textbackslash n" + "b"}; embedding a query with the query prompt versus the passage prompt reorders which document has the $N$th-highest cosine similarity.}

\pitfall{The agent implemented the requested behavior behind an idiomatically-named class, function, or registry key instead of the exact public identifier and signature the task states, so the code builds and runs under its own tests but the checker never reaches it.}
{A hidden checker imports or instantiates a documented identifier (a service class name, a package function, a module registry key) or calls a documented public entrypoint with a stated signature; any differently-named or re-signatured variant is invisible to it even though it behaves correctly in isolation.}
{Before writing code, enumerate the exact public identifiers the prompt names---class name, function name, full parameter list and return type, registry key---and name the artifacts precisely; keep existing public entrypoints and signatures unchanged and add new internals around them rather than renaming the reachable surface.}
{\texttt{from module import ExactName} resolves only when the class/function is literally \texttt{ExactName}; a \texttt{class NearbyName} that subclasses the expected type is not found by an importer that looks up \texttt{ExactName}.}
\end{artifactbox}